%% file: neuraltaskkernels.tex
\documentclass{article}

\usepackage[preprint]{neurips_2025}

\usepackage{hyperref}       
\usepackage{url}            
\usepackage{booktabs}       
\usepackage{amsmath}
\usepackage{amssymb}
\usepackage{amsfonts}       
\usepackage{nicefrac}       
\usepackage{microtype}      
\usepackage{algorithm,algorithmic}
\usepackage{graphicx}
\usepackage[utf8]{inputenc}

\usepackage[T1]{fontenc}
\usepackage{url}
\usepackage{float}
\usepackage{multirow}
\usepackage{tabularx}
\usepackage{titletoc}

\input{math_commands.tex}

\title{Meta-Learning Theory-Informed Inductive\\ Biases using Deep Kernel Gaussian Processes}

\author{%
  Bahti Zakirov \\
  Institute of Science and Technology Austria\\
  \texttt{bzakirov@ist.ac.at} \\
  \And
  Gašper Tkačik \\
  Institute of Science and Technology Austria \\
  \texttt{gtkacik@ist.ac.at}\\
}

\begin{document}

\maketitle

\begin{abstract}

Normative and task-driven theories offer powerful top-down explanations for biological systems, yet the goals of quantitatively arbitrating between competing theories, and utilizing them as inductive biases to improve data-driven fits of real biological datasets are prohibitively laborious, and often impossible. To this end, we introduce a Bayesian meta-learning framework designed to automatically convert raw functional predictions from normative theories into tractable probabilistic models.
We employ adaptive deep kernel Gaussian processes, meta-learning a kernel on synthetic data generated from a normative theory. This \emph{Theory-Informed Kernel} specifies a probabilistic model representing the theory predictions -- usable for both fitting data and rigorously validating the theory. As a demonstration, we apply our framework to the early visual system, using efficient coding as our normative theory. 
We show improved response prediction accuracy in \emph{ex vivo} recordings of mouse retinal ganglion cells stimulated by natural scenes compared to conventional data-driven baselines, while providing well-calibrated uncertainty estimates and interpretable representations. Using exact Bayesian model selection, we also show that our informed kernel can accurately infer the \emph{degree} of theory-match from data, confirming faithful encapsulation of theory structure. This work provides a more general, scalable, and automated approach for integrating theoretical knowledge into data-driven scientific inquiry in neuroscience and beyond.

\end{abstract}

\begin{figure}[h!]
    \centering
    \includegraphics[width=0.8\textwidth]{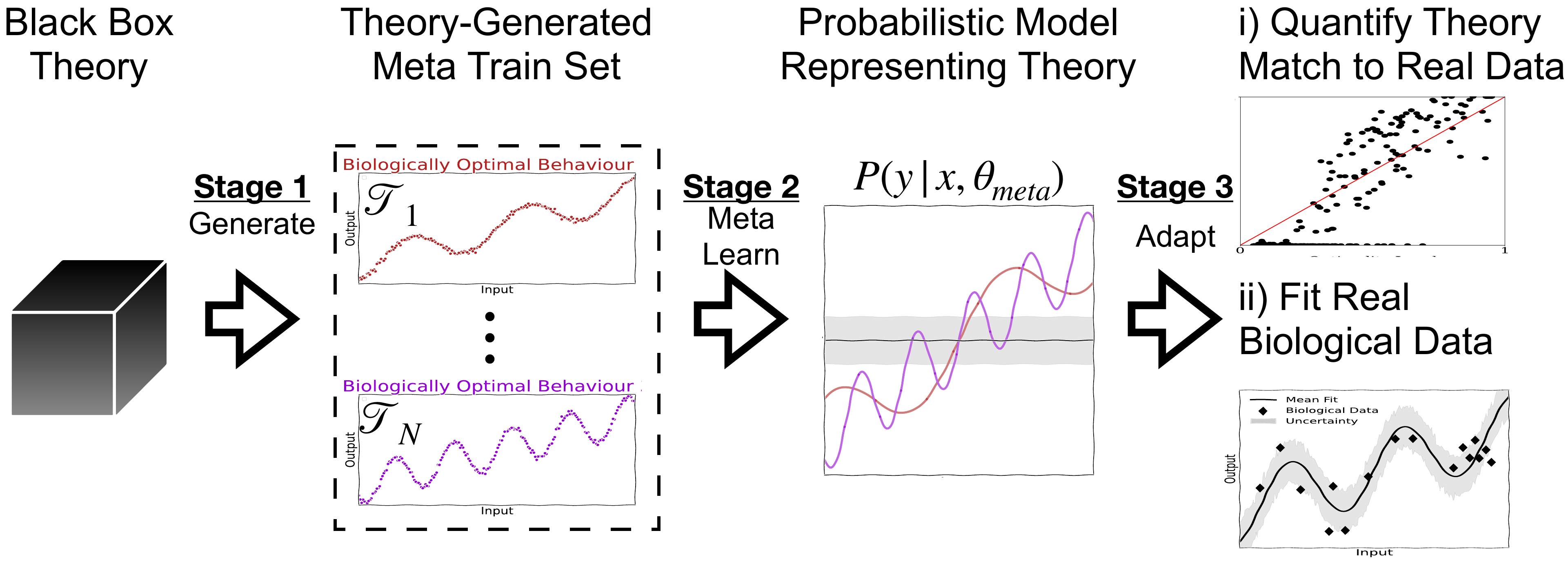}
    \caption{\textbf{Overview of the Proposed Framework}. 
    }
    \label{fig:fig1}
\end{figure}

\section{Introduction}

A striking feature of living matter is \textit{teleonomy}, or apparent goal-directedness. This idea permeates quantitative life sciences as a predictive principle: by assuming organisms evolve under selection for functional phenotypes, we can model their structure and behaviours as optimized solutions to ecologically relevant tasks~\citep{bialek2012biophysics}.
This longstanding tradition, often called \emph{normative}, or more recently \emph{task-driven} modeling, has deep roots in computational neuroscience and sensory coding theory~\citep{barlow1961possible, atick1992could}, having provided a \emph{top-down} explanation for the structure and organization of receptive fields in early visual areas such as the retina and primary visual cortex~\citep{olshausen1996emergence, karklin2011efficient}. It has also proven directly predictive of activity in higher visual, somatosensory, and auditory areas -- often outperforming purely data-driven models~\citep{yamins2014performance,kell2018task,zhuang2021unsupervised,vargas2024task}.

Despite mounting evidence demonstrating the broad predictive and explanatory power of top-down models, this enterprise suffers from two important challenges. 
\textbf{First}, given multiple competing normative theories of the same system, there is no general, principled procedure in neuroscience to quantitatively arbitrate between them.  Solutions like Bayesian Model Selection~\citep{kass1995bayes} do exist, but are challenging to apply due to their requirement of first hand-specifying a probabilistic model representing the theory knowledge, and the typical intractability of the resulting marginal likelihoods, which are difficult to even estimate. \textbf{Second}, despite calls in scientific machine learning for domain-informed inductive biases, it remains difficult to systematically combine the theoretical knowledge of a good top-down model with the noisy, complex reality of empirical data to aid data-driven predictive inference. Current approaches to solving these challenges require laborious expert hand-design, and are either somewhat heuristic and system-specific (for example, constraining the receptive fields of models of early visual systems~\citep{qiu2023efficient,goldin2023scalable}) or limited in applicability to idealized simple models~\citep{mlynarski2021statistical}.

Here, we identify Bayesian meta-learning as a principled methodological solution to both problems. We present a framework designed to automatically construct theory-informed probabilistic models applicable to any system for which top-down theories that can generate input/output predictions are available. We demonstrate our framework using a bespoke variant of adaptive deep kernel meta-learning~\citep{  patacchiola2020bayesian, chen2023meta}: meta-learning a Gaussian process kernel
from synthetic data generated by the well-known~\emph{efficient coding} normative theory of the retina. Concretely, we first meta-learn a deep feature extractor shared across synthetic tasks derived from efficient coding theory. By virtue of meta-learning, this feature extractor forms an abstract metric embedding where the geometric distances between transformed inputs are meaningful for the \emph{entire class} of theory-consistent functions. The frozen embedding is subsequently used and updated by downstream task-adaptive components to form a \emph{Theory-Informed Kernel} for biological data. Our informed kernel defines a Gaussian process prior over the space of functions, or an \emph{inductive bias}, representing the theorized biological purpose of the system. 

We show that this theory-informed inductive bias outperforms conventional data-driven baselines for learning the neural encoding function of \textit{ex vivo} retinal ganglion cells, providing evidence that our framework fruitfully transfers knowledge from theory to real biological data. Further, the probabilistic theory-informed models produced by our framework natively allow the exact computation of the Marginal Likelihoods needed for Bayesian model comparison, thereby enabling rigorous information-theoretic quantification of the match between theory and real data. We show that this is sensitive enough to accurately infer the \emph{degree} of match between theory and data, which is more challenging than a simple binary choice between competing models. All in all, our work provides a much needed bridge between top-down theory-driven understanding, and bottom-up data-driven prediction~\citep{breiman2001statistical}.

\textbf{Our main contribution} is a framework enabling new scientific capabilities for neuroscience that is built upon and contributes to recent advances in Bayesian machine learning. \textbf{For neuroscience}, we introduce a path to automatically construct tractable probabilistic models from ``black-box'' theories (Figs.~\ref{fig:fig1},\ref{fig:fig:2}), enabling the rigorous, information-theoretic validation of competing scientific hypotheses (Fig.~\ref{fig:fig5}) in a way that is insensitive to the theory's parameterization. Our theory-informed model improves predictive accuracy and uncertainty quantification on real data (Fig.~\ref{fig:fig3}), while also providing interpretability and allowing the embedded theory knowledge to be automatically relaxed or enriched as the data demands (Fig.~\ref{fig:fig4}). \textbf{For Bayesian machine learning}, we provide a compelling new application of deep kernels for the open problem of domain-informed kernel design. We demonstrate a successful real-world implementation: introducing a bespoke architecture and other practical design choices useful for future work (\ref{featurecollapse}) that avoid the common ``feature collapse'' pathology~\citep{ober2021promises} to deliver well-calibrated uncertainty (Fig.~\ref{fig:fig3}b,c). Finally, we validate key Bayesian deep learning principles~\citep{wilson2020bayesian} relating generalization to inductive biases (Fig.~\ref{fig:fig3}a) on a challenging scientific problem.

\section{Background}

\textbf{Top-Down Theories in Neuroscience.} \smallskip 
\emph{Normative theories} (also known as optimality theories) propose that the properties of biological systems emerge as a result of evolutionary pressures optimizing a utility functional. In these frameworks, the parameters governing neural coding are regarded as optimal solutions to the underlying ecological tasks. Formally, one may express this as $\theta^* = \arg\max_{\theta}\, U\big[P(r|x,\theta),P_x(x)\big]$

where \(r\) is the neural response, \(x\) denotes the stimulus drawn from a naturalistic distribution \(P_x(x)\), \(\theta\) are the model parameters, and \(U\) a carefully designed utility functional that quantifies how well the response meets the system’s theorized ecological or behavioral goals.

Of particular relevance is the development of efficient coding theories in early sensory systems~\citep{barlow1961possible, atick1992could, olshausen1996emergence,karklin2011efficient,karklin2009emergence,attneave1954some,simoncelli2001natural,lewicki2002efficient, hyvarinen2009natural}. This line of research was among the first to derive the structure of retinal and primary visual cortical encoding by assuming that it optimizes information transmission using limited neural activity. Contemporary \emph{task-driven} extensions operationalize these normative principles by optimizing flexible models, often deep neural networks, directly on ecologically relevant task performance, proving highly predictive of neural activity across various domains~\citep{yamins2014performance, kell2018task, vargas2024task}.

\textbf{Gaussian Processes (GPs).} GPs define distributions over functions such that any finite set of points drawn from the process has a joint Gaussian distribution. In machine learning, GPs serve as Bayesian priors for functions $f$ in regression and classification models~\citep{williams2006gaussian,mackay2003information}. Concretely, we place a GP prior $f_x \sim \mathcal{N}(0, K_{\theta_{\mathrm{gp}}}(\cdot,\cdot))$ with a positive-definite kernel $K_{\theta_{\mathrm{gp}}}$, and link $f_x$ to observed data $y$ through a Gaussian likelihood for regression: $y(x)= \mathcal{N}(f_x,\sigma_\eta I)$. To compute the final predictive equations of all GP's in this work, GP priors are conditioned on training data $\mathcal{D}_{train} = \{X,y\}$ via Bayesian inference: $P(y_{*}|X_{*},\mathcal{D}_{train}) = \mathcal{N} \left( K(X_*, X)[K(X, X) + \sigma_{\eta}^2 I]^{-1}y_, \quad K(X_*, X_*) - K(X_*, X)[K(X, X) + \sigma_{\eta}^2 I]^{-1}K(X, X_*) \right)$, where "$*$" denotes ``test data'', and $K(X_1,X_2)$ is a matrix shaped $|X_1| \times |X_2|$. The kernel entirely specifies the inductive bias, since it determines the mean and covariance functions of the underlying Gaussian prior. \textbf{Deep Kernels} improve GP expressivity by parameterizing the kernel with a neural network ~\citep{wilson2016deep}. Hyperparameters $\theta_{\mathrm{gp}}$ of the kernel and likelihood can be selected by maximizing the marginal likelihood of the data:
\begin{equation}
    \label{mll}
     P(y|x,\theta) \equiv \int P(y|f)p(f|x,\theta)df.
\end{equation}

\textbf{Meta-Learning.}\ \  The goal of meta-learning is to use labeled input/output data from tasks in a \textit{meta-train set} $\mathcal{T} = \{\{x_j,y_j\}_{j=1}^{n_i}\}_{i=1}^{N_{\mathcal{T}}}$ (where $N_\mathcal{T}$ denotes the number of tasks, and $n_i$ the number of data points per task) to learn inductive biases that are useful for fitting on other tasks in the \textit{meta-test set}~\citep{caruana1997multitask,zhang2018overview}. Individual tasks are split into \textit{support} and \textit{query} data (i.e. $n_i = n_i^s + n_i^q$), where the support set is analogous to the training data for that task, and query to the validation data. Meta-learned models may contain task-adaptive components, which (unlike fixed meta-learned components) can be re-fit to every individual task, often on different optimization objectives.

\section{Theory-Informed Meta-Learning Framework}
\label{generalframework}
The framework's application has three steps (Fig.~\ref{fig:fig1}): generating synthetic data from (normative) theory, meta-learning a theory-informed kernel from the synthetic data, and adapting/applying the resulting GP prior to real data.

\begin{figure}[h]
    \centering
    \includegraphics[width=\textwidth]{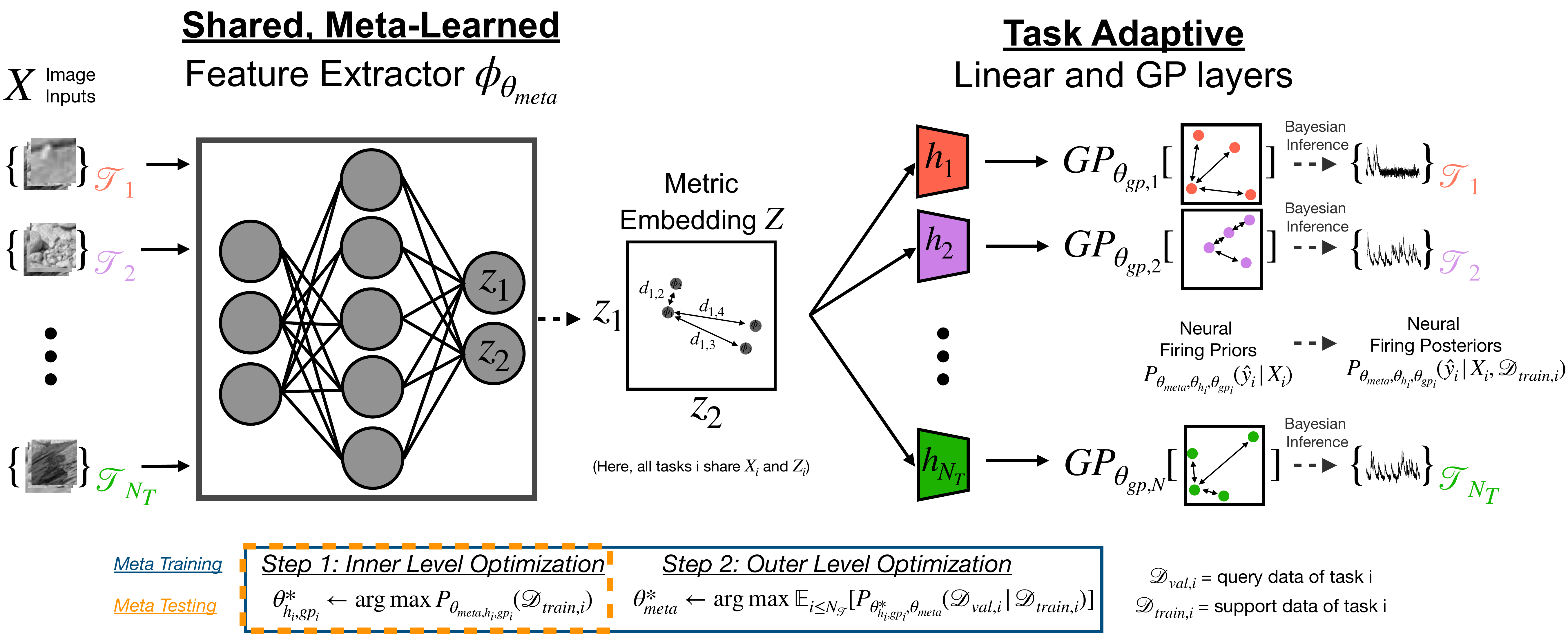}

    \caption{
    \textbf{Task-Adaptive Deep Kernel  Meta-Learning Approach}.
    Inputs (e.g.,\ images) pass through a shared feature extractor $\phi$ whose weights $\theta_{\mathrm{meta}}$ are meta‑learned across tasks.
    The resulting features are passed to a linear head $h_i$ and ultimately the Gaussian process (parameters $\theta_{\mathrm{gp},i}$) layer, both of which are adapted separately to every task $i$ for which we generate predictions (e.g., neural activity). Together, these components constitute the Theory-Informed Kernel. 
    }
    \label{fig:fig:2}
\end{figure}

Our Theory-Informed Kernel comprises disjoint meta-learned and task-adaptive modules (Fig.~\ref{fig:fig:2}):

\begin{itemize}
    \item A meta-learned high-capacity feature extractor $\phi$, shared across tasks, that learns an abstract metric embedding where the geometric distances between inputs are meaningful for the entire class of theory-consistent functions
    \item A task-adaptive linear head (per task) , $\{h_i\}_{i=1}^{\text{N}_\mathcal{T}}$, that adapts the initial metric embedding, mitigating negative transfer~\citep{standley2020tasks} and bridging the gap between simulated and real data,
    \item A task-adaptive Gaussian process layer (per task) that makes predictions by conditioning the theory-informed prior on the transformed inputs. Here we use the \emph{Radial Basis Functions} kernel $K_{\text{RBF}}(z,z';\theta_{\text{gp}})= \sigma_f\exp\left(-\frac{|z-z'|^2}{2\ell^2}\right)$, with hyperparameters $\theta_{\text{gp}}=\{(\sigma_f,\ell,\sigma_\eta)_i\}_{i=1}^{\text{N}_\mathcal{T}}$ (denoting the output scale, length scale, and likelihood noise level respectively).
\end{itemize}
The meta-training procedure (Algorithm \ref{alg:adaptive-deep-kernel}) employs bi-level optimization akin to~\citep{chen2023meta}: first an inner loop fits the task-specific linear head ($h_i$) and GP hyperparameters ($\theta_{\text{gp},i}$) using support data for each task by optimizing Eq. \ref{mll}, then an outer loop updates the shared feature extractor ($\phi$) based on the adapted model's log-likelihood on query data.  Once trained, $\phi$ is frozen, while task-adaptive modules freshly adapt the prior to any meta-test task (\ref{sec:taskadaptationprocedure}, Eq \ref{mll}). \textbf{The final Theory-Informed Kernel} (TIK) specifying the prior for each task $i$ is thus $K_{\text{TIK},i}(x,x')= \sigma_{f,i}\exp\left(-\frac{|h_i(\phi(x))-h_i(\phi(x'))|^2}{2\ell_i^2}\right)$.

\section{Example: Theory-Informed Kernel for Efficient Coding in the Retina}
\label{example:efficientsensorycodes}

We demonstrate our framework on mouse retinal ganglion cell (RGC) responses to natural images.

\subsection{Biological Data and Preprocessing}

We gratefully use previously published \textit{ex vivo} calcium imaging data from 86 RGCs of a single mouse retina responding to 50-frame natural movie sequences presented at 30Hz in UV channels~\citep{qiu2023dataset}. Inputs are derived from downsampled $36\times32$ pixel movie frames. A single measured fluorescence response  (output $\in \mathbb{R}$) is recorded per 50 preceding movie frames for each neuron, as a proxy for that neuron's activity (\ref{sec: biological_data}). Each neuron's responses to all stimuli constitute a separate task, and we use the terms \textit{neurons} and \textit{tasks} interchangeably throughout this section. The dataset came pre-split (16200/750 train/test+val) and prepreprocessed. To adapt the data for image-based models, we temporally flattened the movies using a canonical temporal filter derived from Linear-Nonlinear (LN) model fits to the neural data (\ref{sec: biological_LN_filter}). We then eliminated near-duplicate images via hierarchical clustering (\ref{sec: biological_removing_duplicate}), yielding (1452/(400,350) train/(test,val)) unique images $\in \mathbb{R}^{36\times32}$.
We therefore have \textbf{86 biological meta-test tasks}, each corresponding to the responses ($\in \mathbb{R}$) of one biological neuron to $1452$ natural images $\in \mathbb{R}^{36\times32}$. 

\subsection{Constructing the Theory-Informed Meta-Train Set}
\label{syntheticmetatrain}

To synthesize the meta-train set, we adapted the efficient coding (EC) normative model of Ocko et al.~\citep{ocko2018emergence}. This  model is a convolutional autoencoder trained on natural images with the EC objective of maximizing reconstruction accuracy while minimizing bottleneck activity given constraints. A single synthetic meta-train task, $T_i$, corresponds to learning the linearized encoding function of a single optimized bottleneck neuron. Given a set of N natural images ($x_k$, $36\times32$ pixels), each task is to predict the $N$ corresponding scalar responses, defined by the dot product $y_{i,k} = f_i^Tx_k$, where $f_i$ is the neuron's specific receptive field (RF). The linearization isolates the well-validated~\citep{hyvarinen2009natural,karklin2011efficient} center-surround RFs that emerge as the single-neuron consequences of the population-level efficient coding objective.  We extracted these RFs ($f_i$) from the optimized bottleneck by fitting an LN model to its activations. We then applied standard data augmentation and preprocessing to extend the meta-train set, resulting in a final \textbf{490 synthetic meta-train tasks derived from EC theory}. Full details of this procedure, including rationale and sensitivity analysis, are provided in \ref{app:synthetic_data}

\begin{figure}[tb]
    \centering
    \includegraphics[width=0.8\textwidth]{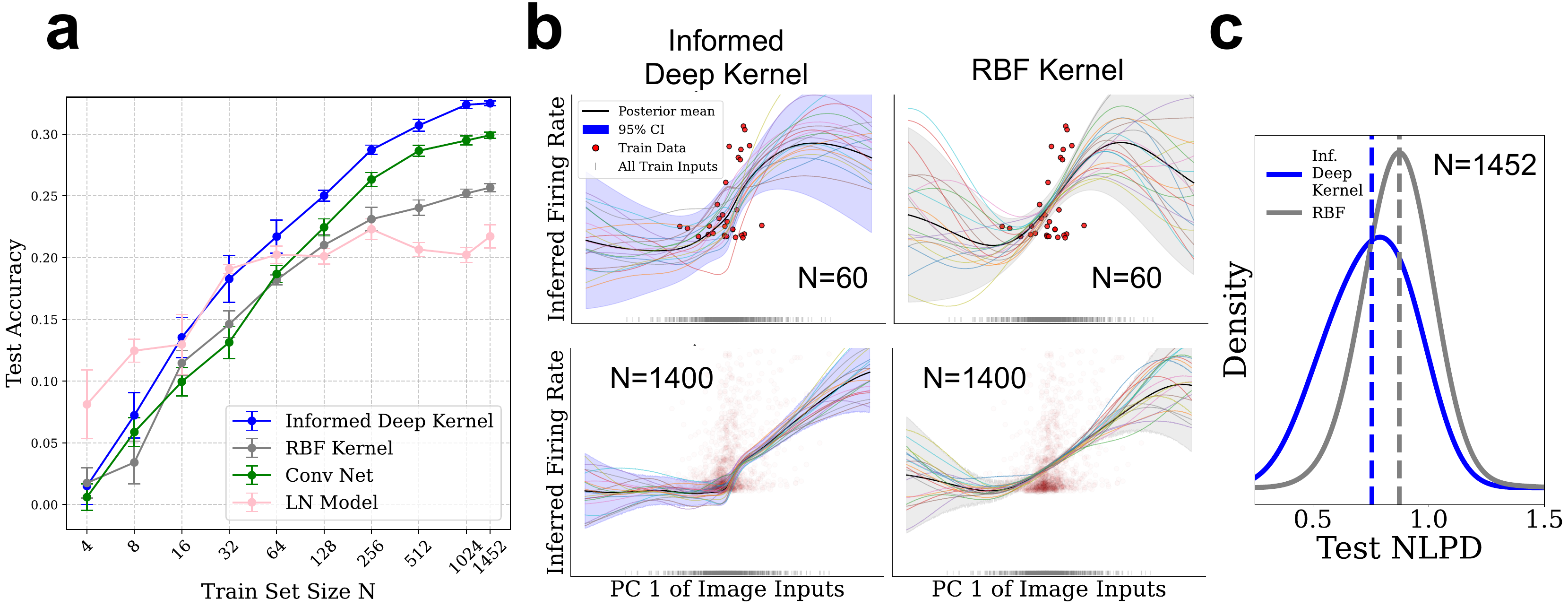}
    \caption{\textbf{Theory-Informed Kernel improves fit on retinal data.}
    \textbf{(a)}  Test accuracy of model predictions at varying train set sizes $N$ (mean Pearson correlation over 86 neurons, standard error over 5 seeds).
    \textbf{(b)} Posterior mean (solid line), epistemic uncertainty (shaded), posterior samples (colored lines) of tested GP models over the first principal component (PC1) of train images. Models conditioned on $N=60$ (top) and $N=1400$ (bottom) points (red dots). Projected full training images shown as gray ticks (above x-axes). 
    \textbf{(c)} Kernel Density Estimate (KDE) of the NLPD scores ; lower is better) of the RBF and the Theory-Informed kernels (all 86 neurons, 5 seeds).}
    \label{fig:fig3}
\end{figure}

\subsection{Theory-Informed Kernel Improves Learning}
\label{result1}

Following Sec.~\ref{generalframework}, we meta-learn the feature extractor ($\phi$) on our theory-generated data~(Sec.~\ref{syntheticmetatrain}). Freezing $\phi$, we fit the task-adaptive components ($h_i$ and $\theta_{\mathrm{gp},i}$) for each neuron $i$ ($N_{\mathcal{T}}^{\text{test}}$= $86$) using Eq.~\ref{mll}. To assess the transfer of our meta-learned inductive bias, we then compare performance metrics against standard neuroscience baselines. Our first baseline is the well-validated \emph{Systems Identification} convolutional neural network (CNN), specifically designed for this task~\citep{qiu2023efficient}. Our second baseline is the standard \emph{Radial Basis Functions} (RBF) GP, with kernel parameters set by optimizing Eq.~\ref{mll}. We also compare against the Linear-Nonlinear (LN) model, transfer learning, and ablated versions of our own model,  designed to ensure that the transferred features reflect knowledge obtained from the theory (\ref{sec: ablations}). We quantified the test accuracy using the Pearson correlation between predicted and measured neural responses. Hyperparameters were identified via cross validation; early  stopping was used for our validation data (350 samples), equally for all models (\ref{sec:baselinedetails}).

Fig.~\ref{fig:fig3}a depicts the performance of our model alongside  the baselines, as a function of the number of training data points $N$.

The experimental results follow clear trends, with the RBF GP performing competitively with the CNN at low data sizes ($N\leq64$), and the LN outperforming the Theory-Informed Kernel in extremely low data regimes, where it is difficult to learn even the linear heads ($N\leq 8$). Nonetheless, across a large range in $N$, we find that our model outperforms all baselines (Fig.~\ref{fig:fig3}a, \ref{sec: ablations}). Because our approach ultimately feeds the data embeddings obtained by the meta-learned feature extractor $\phi$ and linear heads $h_i$ into its own RBF GP, the improved performance over the bare RBF GP indicates that the learned embedding is meaningful. Because the Theory-Informed Kernel significantly outperforms ablated versions where we either randomize or remove the meta-learned $\phi$ (\ref{sec: ablations}) and the meta-train set, we conclude that $\phi$ indeed encapsulates knowledge from the efficient coding theory, thereby improving performance by a higher margin than learning the task-adaptive parameters alone, or simply meta-learning on a generic task would allow.

Comparatively, our model achieves both high data efficiency for small $N$, as well as the highest accuracy using the large $N$ (full dataset). This subverts the conventional idea about kernel methods, such as  GPs, being specialized for extremely low data regimes; instead, it strongly supports the contemporary narrative in probabilistic deep learning, suggesting that well-calibrated but flexible inductive biases are key to powerful models that perform well with any amount of data~\citep{wilson2020bayesian}.

\subsection{Theory-Informed Kernel Preserves Epistemic Uncertainty}
\label{uncertainty}

A crucial property of probabilistic methods like ours is their ability to represent uncertainty in their predictions. In uncertainty quantification (UQ), an important distinction is made between \emph{epistemic} and \emph{aleatoric} uncertainties~\citep{hullermeier2021aleatoric}, with the former being unpredictability reducible by observing more data or improving model specification, and the latter being irreducible unpredictability (e.g.,  noise). Due to pathologies that arise from the interaction between deep learning and the marginal likelihood objective (Eq.~\ref{mll}), Deep Kernel GPs notoriously tend to exhibit \emph{feature collapse}, specifically resulting in overconfident epistemic uncertainty~\citep{van2021feature,liu2020simple,ober2021promises}. Naturally, this is a concern for our framework. 

To assess whether our model's representation supports reasonable uncertainty quantification, we compare it to the RBF GP, with both models fit on the full dataset. We first present, on the best fit neuron for both models ($i=37$), visualizations of their predictions and 95\% CI (here only epistemic uncertainty), as a function of the first principal component (PC1) of the full stimulus dataset (fixing the other PCs to their means). We show cases where both models are conditioned on either $N=60$ or $N=1400$ data (Fig.~\ref{fig:fig3}b). Both models exhibit the characteristic reduction of uncertainty near the observed data at $N=60$, but unlike the RBF GP, our Theory-Informed Kernel  heteroskedastically preserves some uncertainty near $\text{PC1}=0$ where the samples are dense. Conditioned on $N=1400$ data, the epistemic uncertainty of the RBF GP collapses near the data, while the informed kernel maintains a reasonable confidence region enveloping a substantial proportion of the data (despite only showing the epistemic part). This suggests, qualitatively, our representation usefully structures the input space, reducing overconfidence in regions with many observations. 

Next, to quantitatively assess UQ in our model, we compute the standard Negative Log Predictive Density (NLPD) metric on the full images: $\text{NLPD}(y_{\text{test},i}|\text{model}_i)\equiv - \log P(y_{\text{test,i}}|x_{\text{test},i},x_{\text{train},i},y_{\text{train},i},\phi,\theta_{\text{gp},i},h_i)$

Fig.~\ref{fig:fig3}c shows the kernel density estimate of NLPD values across all neurons and data. Our Theory-Informed Kernel exceeds the UQ capabilities of the RBF baseline on biological data, avoiding common Deep Kernel GP pitfalls~\citep{ober2021promises} and enabling potential downstream applications where it is vital for the model to ``know what it doesn't know''~\citep{cowley2017adaptive,frazier2018tutorial}.

\begin{figure}[h]
    \centering
    \includegraphics[width=0.7\textwidth]{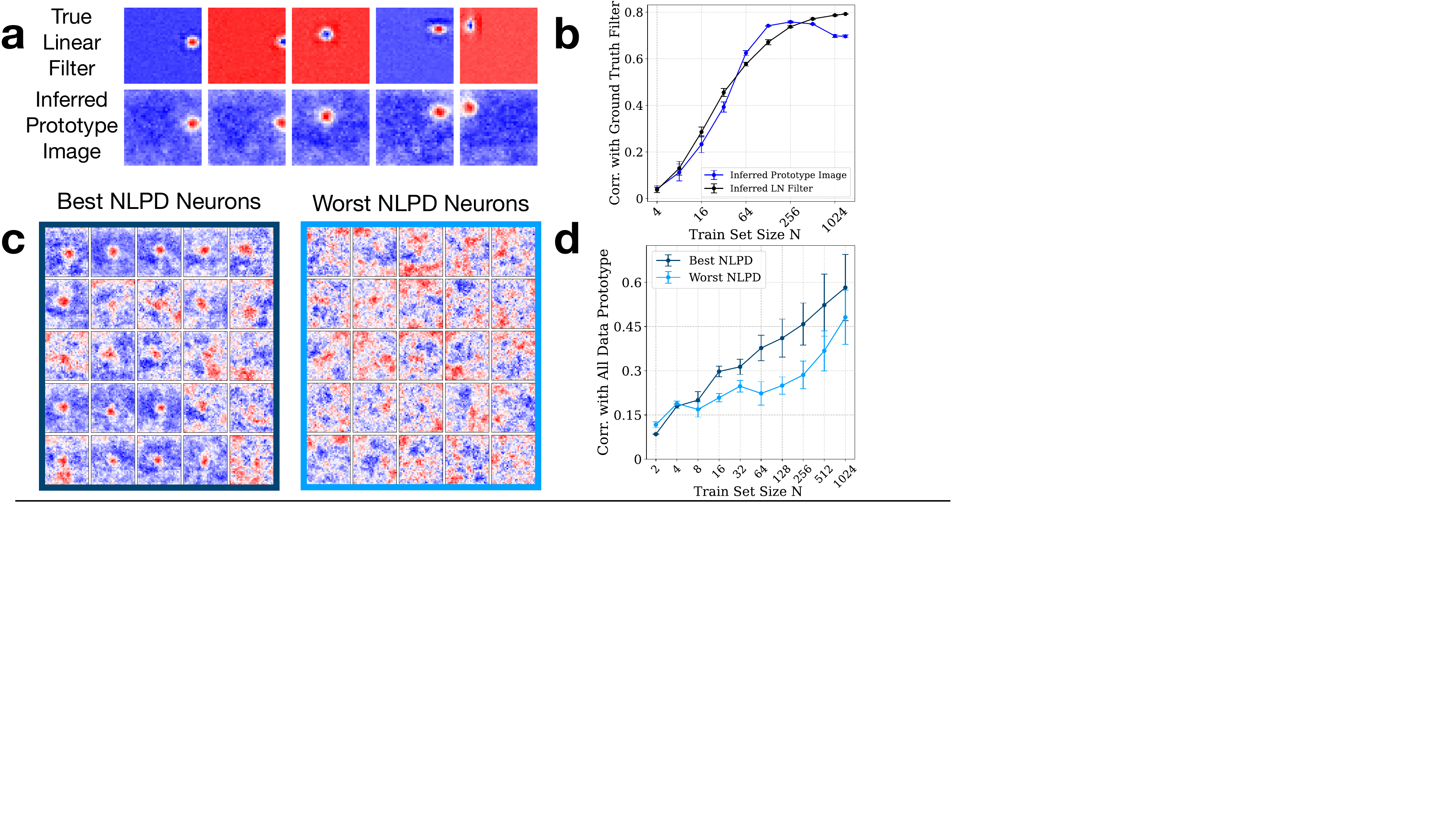}
    \caption{\textbf{Task-Adaptive Heads Learn Interpretable Representations.} 
    \textbf{(a)} Prototype images ($\mathcal{P}_i$) derived from learned representations vs. ground truth linear filters of synthetic neurons.
    \textbf{(b)} Mean Pearson corr. between the absolute values of the $\mathcal{P}_i$ (blue) and filters learned by an LN model (black) with ground truth as in (a), varying train set size $N$. 
    \textbf{(c)} $\mathcal{P}_i$ for the 25 best (dark blue) and worst (light blue) performing real neurons, sorted by NLPD, learned from the full dataset ($N=1452$).
    \textbf{(d)} Mean Pearson corr. between $\mathcal{P}_i$ at varying $N$, with the $\mathcal{P}_i$ at full dataset ($N=1452$), split according to NLPD as in (c). (b) and (d) averages over neurons with standard errors across 5 random seeds.
    }
    \label{fig:fig4}
\end{figure}

\subsection{Theory-Informed Kernel Learns Interpretable Representations}
\label{interpretability}

How exactly does our model's learned representation support its predictive performance? As a Deep Kernel GP with an RBF layer, its predictions are driven by the pairwise Euclidean distances between transformed inputs.
To gain mechanistic insight into how the task-specific linear transformations $h_i$ tailor the model's function, we analyze their impact on these critical pairwise distances.  Starting with a set of probe images $\pi$, we compute a \emph{prototype image}, $\mathcal{P}_i$, for each $h_i$, by summing the pixel-level overlap masks between pairs of images $x_m,x_n \in \pi$ weighted by how $h_i$ alters their projected distances $\lVert h_i(\phi(x_m)) - h_i(\phi(x_n)) \rVert_2$ (\ref{app:prototype_derivations}). These $\mathcal{P}_i$ represent features of the inputs that $h_i$ learn to `pull together' or `push apart' for their task $i$, in their output representation distances.

We first inspect the prototype images on synthetic data, finding that prototype images $\mathcal{P}_i$ recapitulate the filters used to generate the theory-informed responses very well (Fig.~\ref{fig:fig4}a). We further observe that the rate at which the prototype image of the Theory-Informed Kernel recovers (in absolute value) the ground truth RF filter as a function of train set size $N$ closely matches the rate of recovery by the LN model's linear filter (Fig.~\ref{fig:fig4}b), indicating the importance of  $\mathcal{P}_i$  for this model. At high train set size $N$, however, the correlation between the ground truth and prototype image stops increasing monotonically, in a textbook example of the \emph{Bayesian Occam effect}~\citep{rasmussen2000occam,mackay1992bayesian,mackay2003information}.  As $N$ grows, the GP layer’s non‑parametric capacity automatically expands, so the marginal likelihood stops rewarding additional structure in $h_i$ and favors a less informed prior, letting the data ``speak for themselves''. This contextualizes our argument from Sec.~\ref{result1} about  \emph{informed yet flexible} models that  retain their advantage across all dataset sizes: when data are scarce, the theory‑informed inductive bias supplies the needed structure; when data are plentiful, Eq.~\ref{mll}  automatically relaxes that structure to permit a predictive fit.

Applied to biological data, we find a clear distinction between $\mathcal{P}_i$ corresponding to tasks $i$ on which the model performed well, and those on which it performed poorly. The 25 best performing (by NLPD) $h_i$ transform the data in interpretable ways: resembling what may be the receptive fields of the biological neurons (Fig.~\ref{fig:fig4}c). The 25 worst $h_i$, conversely, seemingly fail to do so. This is reflected in (Fig.~\ref{fig:fig4}d), which depicts the Pearson correlation between the prototype images learned at various $N$, with the prototype image given the full data; the worst performing tasks learn a less interpretable and also less consistent prototype image.  Unlike the synthetic case, for real neurons the structure in $\mathcal{P}_i$  keeps increasing up to $N=1024$, indicating that we are still in a prior‑aided regime where the data alone have not saturated the non‑parametric capacity.

Taken together, our Theory-Informed Kernel framework is both \emph{performant} and \emph{interpretable}. When it fails, we gain insight into why; when it succeeds, we can watch the Occam effect unfold, demonstrating that Bayesian principles can deliver practical gains in real‑world settings.  
Moreover, our prototype‑based interpretation is readily transferable to domains beyond sensory neuroscience, wherever one wishes to visualize what a deep‑kernel GP has learned to look for.

\begin{figure}[h]
    \centering
    \includegraphics[width=0.8\textwidth]{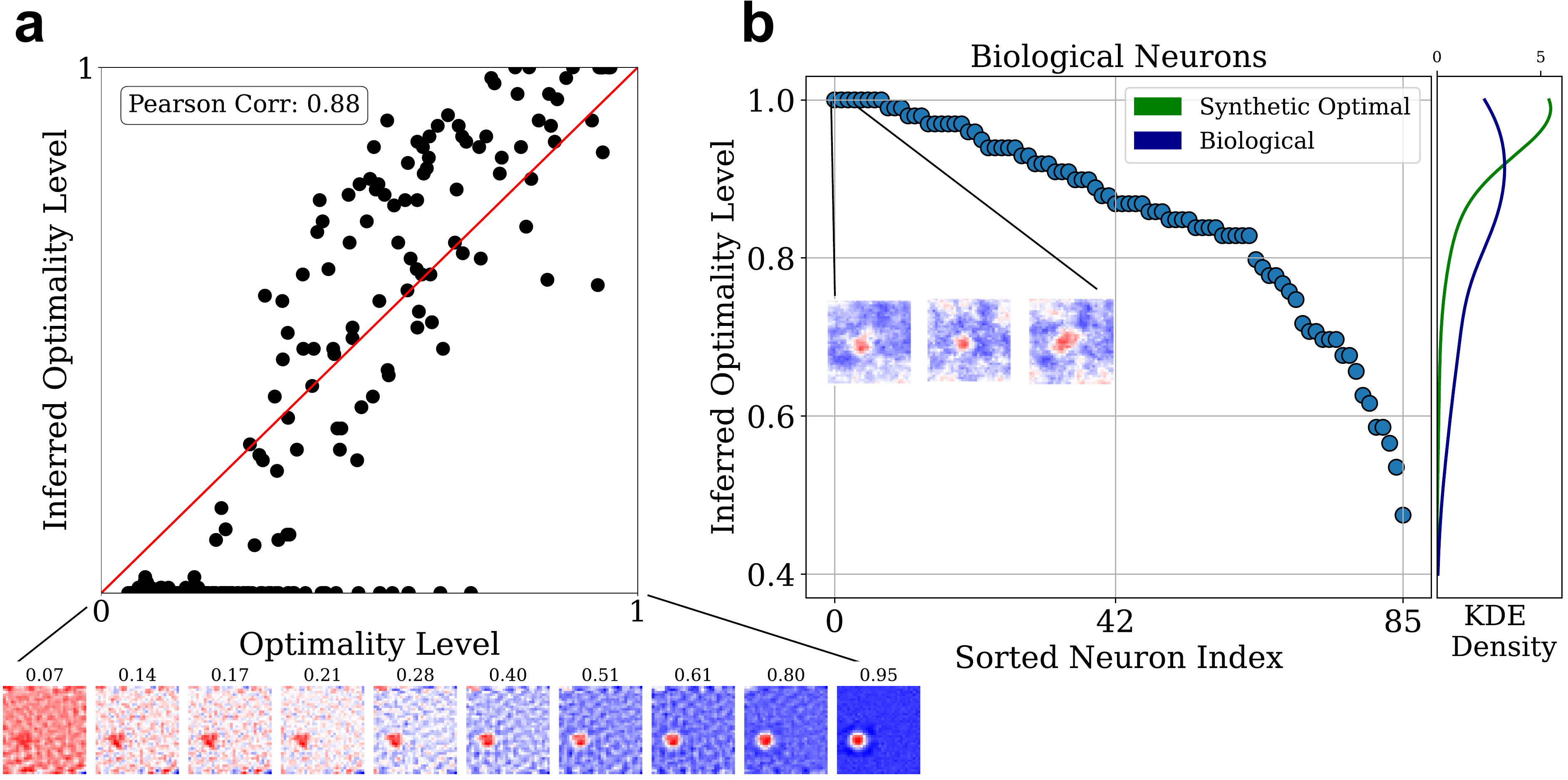}
    \caption{\textbf{Theory Informed Kernel quantifies Theory Fidelity.} 
    \textbf{(a)} Degree of optimality, $\beta^*$, inferred by our framework, plotted against ground truth on synthetic neurons (black dots; Pearson correlation $=0.88$).  Subpanel: example synthetic receptive fields of varying optimality.
    \textbf{(b)} Scree plot with the inferred efficient coding level of optimality $\beta^*$ for all 86 recorded retinal ganglion cells. Inset: ``Prototype images'' $\mathcal{P}$ of top three RGCs (detailed in Fig.~\ref{fig:fig4}). Right margin: KDE of inferred optimality for all 86 RGCs (blue) and 100 synthetic optimal neurons (green).}

    \label{fig:fig5}
\end{figure}

\subsection{Theory-Informed Kernel Enables Validation of Black Box Normative Theories}
\label{result3}
A major methodological challenge in top-down (either normative or task-driven) modelling is the validation of how well a hypothesized optimality theory describes the biological system.  Here, we cast the problem of quantifying the degree to which an optimality theory describes the real dataset as \emph{Bayesian Model Comparison} (BMC). Typically, BMC involves comparing the marginal likelihood, $P(\mathcal{D}|\mathcal{M})$, which quantifies the evidence for competing hypotheses (models $\mathcal{M}$) given the data $\mathcal{D}$. A Bayesian might decide between $\mathcal{M}=1$ (e.g., a specific theory) and $\mathcal{M}=0$ (e.g., a null hypothesis) by selecting the highest scoring $\mathcal{M}$. In our GP framework, this suggests a comparison between the theory-informed kernel and a strong yet theory-agnostic \emph{null model} kernel like the RBF. However, a simple binary choice between these two alternatives may be inadequate for biological systems, which often only partially adhere to the postulated theoretical principles (e.g., neurons sometimes exhibit  deviations from the theoretically-predicted center-surround receptive fields). We thus propose $K_\beta$, interpolating between  theory-informed (TIK) and theory-agnostic (RBF) kernels:
\begin{equation}
    K_\beta(x,x') = \beta K_{\text{TIK}}(x,x') + (1-\beta)K_{\text{RBF}}(x,x')
\end{equation}

Since the marginal likelihood quantifies the evidence for the specific prior defined by $K_\beta$, the optimal  $\beta^* = \arg \max _\beta P(Y|X,\beta)$ indicates the degree to which the data provides evidence for the theory-informed structure beyond what can be captured by the powerful-yet-generic null kernel. In other words, $\beta^*$ can be interpreted as the inferred measure of the theory's relevance, or alternatively, the system's \emph{degree of postulated optimality}.

We first validate this approach on synthetic tasks. We generate these by starting with 20 optimal synthetic receptive fields (RFs) unseen during meta-training, and gradually add structured noise to push them away from optimality (\ref{app:generating_suboptimal}). As a proxy for the ground truth optimality of these new synthetic RFs, we use the standard procedure of computing the $R^2$ of a fit of the \emph{Difference-of-Gaussians} (DoG) model to the RF, as DoG are theorized to constitute the efficient coding solution for the retina (\ref{app:dog_fitting})~\citep{atick1992does}. We then freeze the feature extractor $\phi$ and fit the task-adaptive parameters (linear heads $h_i$ and GP parameters $\theta_{\mathrm{GP},i}$) for $K_{\text{TIK}}$, as well as the parameters $\theta^{\mathrm{null}}_{\mathrm{GP},i}$ of $K_{\text{RBF}}$, separately on each task $i$. Finally, with all parameters frozen for both kernels, we find optimal $\beta^*_i = \arg\max_{\beta \in [0,1]}{\text{P}(Y_i|X_i,K_\beta)}$ via grid search over 100 points. 
%
%
Fig.~\ref{fig:fig5}a shows that $\beta^*$, inferred neuron by neuron as described, correlates strongly with the ground truth optimality level on synthetic data.  Given a postulated normative theory, $\beta^*$ therefore approximates the neuron's \emph{degree of optimality}. We validate the importance of our theory-informed features with an ablation study in \ref{sec:bmc_ablations}.

Next, we repeat the same procedure on our biological RGC dataset, by estimating $\beta^*_i$ for each neuron $i$. The inferred optimality levels of all $86$  neurons are presented in Fig.~\ref{fig:fig5}b, as an example of  a scientifically relevant analysis enabled by our Bayesian framework. Most neurons score highly on the proposed optimality metric $\beta^*$, consistent with the general consensus about the applicability of the efficient coding theory~\citep{simoncelli2001natural,pitkow2012decorrelation}. For comparison, Fig.~\ref{fig:fig5}b also depicts, in the right margin, the $\beta^*$ of 100 randomly selected synthetic optimal neurons (unseen in meta-training). 

The ability of our framework to quantify how well a top-down theory  explains neural data is a practical step forward for neuroscience. Though BMC has long been advocated as the principled solution to this problem~\citep{jaynes2003probability}, its routine use has been hindered by two obstacles: (i) manually translating a normative theory into a predictive probabilistic model; and (ii) computing the resulting marginal likelihoods in high‑dimensional neural datasets. Consequently, most studies still rely on heuristic comparisons of summary statistics, such as the receptive field structure or tuning curve parameters~\citep{ringach2002orientation,karklin2011efficient}. By embedding normative theories into a Gaussian process framework and leveraging the exact tractability of Gaussian marginal likelihoods, our approach addresses both obstacles. This makes principled Bayesian model comparison more feasible at scale and lowers the barrier for quantitatively testing normative theories on neural data.

\section{Discussion and Conclusion}

\textbf{Related Work.}\ \ 
Concepts of meta-learning and domain-informed inductive biases pervade science, and scientific machine learning~\citep{karniadakis2021physics,lake2017building,mccoy2020universal}.
Our framework was inspired by \citet{linderman2017using}'s call to embed  high-level computational goals of neural systems into statistical models. In line with their Bayesian tenets, where theory (via priors) is combined with data (via likelihoods),  our work delivers on their promises: enhancing interpretability and identifiability, and enabling empirical validation of theories.
We also build on the elegant framework of \citet{mlynarski2021statistical}, who incorporate maximum entropy priors derived from top-down theories into parametric Bayesian models, thereby improving inference and validating optimality theories using BMC. Similarly, \citet{bittner2021interrogating} use normalizing flows to represent the distribution of mechanistic model parameters supporting desired emergent computational properties. In contrast, our approach sidesteps the need to commit to any specific parametric form, by meta-learning the priors directly in function space, yielding increased generality and scalability without compromising Bayesian principles. 
We are indebted to the work of  \citet{qiu2023efficient}, who use efficient coding as an inductive bias for predictive modelling on their RGC dataset which we use here as well. Their approach hinges on enforcing shared spatial filters for two CNN models trained in parallel: one performing efficient coding (EC), the other fitting data, yielding improved accuracy and biologically plausible CNN filters. Unlike Qiu et al, who specifically design an inductive bias to exploit EC's ability to produce useful spatial filters for a CNN, our framework does not assume the biological system to be modeled or the theory knowledge to be meta-learned, thus promising applicability to a broader class of systems. Finally, our GP architecture is a variant of \citet{chen2023meta} meta-learner, to which we have introduced the $h_i$ (crucial to bridge the sim-to-real gap), and adjusted the training to avoid feature collapse (\ref{featurecollapse}). Architecture aside, our work fundamentally differs: focusing on theory to data transfer rather than data-driven meta-learning. We elaborate in (\ref{sec: extended_related_work}), particularly contrasting our work with that of Qiu and M\l ynarski.

\textbf{Weaknesses and Future Directions.}\ \ 
\textbf{(i)} A way to inspect \emph{what} meta-learned knowledge is actually stored in the shared feature extractor $\phi$ would be very useful, especially for domains other than efficient coding where predictive aspects of normative theories are less well  understood.  Although this is a generic limitation of meta-learning, the inherent interpretability of deep kernel representations, demonstrated in Sec.~\ref{interpretability}, may provide a stepping stone for future work.
\textbf{(ii)} In seeking a principled separation between meta-learned knowledge and task-specific adaptations, we deliberately used a frozen, meta-learned feature extractor $\phi$ in our kernel. 
An exciting open question is whether the knowledge in $\phi$ could be enriched by carefully allowing its adaptation also on biological data. \textbf{(iii)} Predictions relating to Bayesian Model Comparison suffer from all of the commonly known challenges of using BMC with powerful models~\citep{kass1995bayes,gelman1995bayesian,bernardo2009bayesian}: namely, sensitivity to data properties, to the null model, and to the adaptation of the prior. Finally, \textbf{(iv)}, though our framework is by design general, our empirical validation is limited to a single challenging system. We justify our claim to generality in (\ref{sec: generality}).

While there are many prospective improvements that are interesting research directions, we view them as promising future extensions to our framework and not as refutations of its contributions.

\textbf{Conclusion.} \ \ 
We have presented a Bayesian framework to meta-learn flexible probabilistic models from top-down theories. Using deep kernel GPs, we've successfully transferred knowledge from the efficient coding theory of early visual processing to model \emph{ex vivo} neural recordings, demonstrating interpretability and improving accuracy, uncertainty quantification, and data efficiency compared to baselines and ablations. Through Bayesian model comparison, our framework provides a new scalable path to quantitatively validate normative theories against empirical data—an advancement with far-reaching potential for neuroscience. 
\newpage

\subsection{Acknowledgements}
This research was funded in whole or in part by the Austrian Science Fund (FWF) 10.55776/COE16. We thank Wiktor Młynarski and Henry Moss for helpful conversations. We thank Max Cairney-Leeming and James Odgers for proofreading the manuscript.

\bibliographystyle{iclr2026_conference}
\bibliography{sn-bibliography.bib}


\newpage
\appendix 

\setcounter{page}{1}
\setcounter{section}{0} 
\setcounter{table}{0}
\setcounter{figure}{0}
\setcounter{equation}{0}

\renewcommand{\thepage}{S\arabic{page}}
\renewcommand{\thesection}{S.\Alph{section}}

\renewcommand{\thetable}{S\arabic{table}}   %
\renewcommand{\thefigure}{S\arabic{figure}} %
\renewcommand{\theequation}{S\arabic{equation}} %

\renewcommand{\theHfigure}{S\arabic{figure}}
\renewcommand{\theHtable}{S\arabic{table}}
\renewcommand{\theHequation}{S\arabic{equation}}

\section*{Supplementary Material: Meta Learning Theory Informed Inductive Biases using Deep Kernel Gaussian Processes}

\addcontentsline{toc}{section}{Supplementary Material} 
\startcontents[supplement] 
\titlecontents{section}      
  [3.8em]                     
  {\sffamily}                 
  {\contentslabel{3em}}       
  {\hspace*{-3em}}            
  {\titlerule*[1pc]{.}\contentspage} 

\titlecontents{subsection}
  [7.0em] 
  {\sffamily}
  {\contentslabel{3.2em}} 
  {\hspace*{-3.2em}}
  {\titlerule*[1pc]{.}\contentspage}

\input{suppmat_content.tex}

\end{document}

%% file: math_commands.tex
\usepackage{amsmath,amsfonts,bm}

\def\eqref#1{equation~\ref{#1}}

\def\1{\bm{1}}

\DeclareMathAlphabet{\mathsfit}{\encodingdefault}{\sfdefault}{m}{sl}
\SetMathAlphabet{\mathsfit}{bold}{\encodingdefault}{\sfdefault}{bx}{n}



%% file: suppmat_content.tex
\printcontents[supplement]{}{1}{\setcounter{tocdepth}{2}}

\newpage
\section{Extended Related Work}
\label{sec: extended_related_work}
Here we extend the discussion on some closely related work, showing how our framework differs.

\cite{mlynarski2021statistical} present work related to ours- a framework to construct normative-theory informed parameter priors for parametric models. They specify their normative prior numerically using a specific utility function $U(\theta)$ for a hand-crafted parameterization in $\theta$. Their prior, $\frac{\exp[\beta U(\theta)]}{Z(\beta)}$ crucially requires computing a normalization constant $Z$ by integration, which they estimate by monte carlo but, as they note, is intractable for even ~1000 dimensional cases (e.g. our RGC case). Their model comparison application also hinges on this constant (equiv. to our marginal likelihood (ML)), and therefore must be approximated. These intractabilities limit the applicability of Młynarski et al's framework to small, simple models with well understood, well parameterized normative theories. In contrast, our framework works with any theory that can generate raw predictions (e.g. task-driven networks, autoencoders), as it learns the prior in function space, not parameter space. We are not constrained by dimensionality, and compute our ML analytically. Our framework thus is more automated, general, and scalable while being similarly principled and rigorous.

While we build on the work of \cite{qiu2023efficient} and have benefited greatly from their contributions, our work fundamentally differs from theirs in important ways: \textbf{First}, our model is probabilistic, providing well-calibrated uncertainty estimates (Fig. \ref{fig:fig3}c). Furthermore, our normative inductive bias is adaptive to new tasks, and as we show in our analysis of the Occam's Razor effect (Fig. \ref{fig:fig4}b), can even relax its influence as more data becomes available. \textbf{Additionally}, the model of Qiu et al is well designed to support one theory. Their method hard-codes constraints informed by the theory into the first-layer convolutional filters, limiting it to Retinal ganglion cell applications. Our framework's function-space prior is much more general, allowing it to model abstract knowledge that cannot be encoded in a first-layer spatial filter (As discussed in \ref{sec: generality}). \textbf{Finally}, through its use of exact marginal likelihoods, our framework provides a principled, information-theoretic tool to quantitatively arbitrate between competing and structurally different theories. The closest thing Qiu et al's model can do is enable quantitative comparison of learned receptive fields- which, while valuable, is only a useful heuristic limited in applicability to this one system. 

\paragraph{Other related works}
The work of \cite{shrinivasan2023taking} shares our goal of rigorously testing normative theories with real data. However, our frameworks differ in their core approach and capabilities. Shrinivasan et al focus on manually instantiating variants of a single normative theory (the Neural Sampling Code (NSC)) as probabilistic models. They instantiate the whole generative model hypothesized by NSC $P(x,z)$ and test various competing parameterizations for the constituent $P(z|x)$ and $P(x)$, maximizing the joint likelihoods. Our work is similar in that we also principled instantiate model comparison on the basis of likelihoods of probabilistic models. In contrast to their work, we do not hand-parameterize and then fit a normative-theory inspired probabilistic model, but rather we meta-learn the probabilistic model, represented by the GP kernel function. Thus, the goal of our framework is to enable analysis like that of Shrinivasan et al but for a broader class of theories that may be difficult to hand-parameterize as a probabilistic model directly. The scientific value of Shrinivasan's work substantiates the need for a framework streamlining the process, like ours. Methodologically, ~\cite{moss2024deep} also fit deep kernel GPs on simulated data, but focus on learning a solution operator to speed up latent force models, rather than a theory-informed kernel for data modeling. Our work is also related to that of \cite{kobalczyk2025towards}, who also use Bayesian meta-learning to incorporate human-interpretable domain knowledge into probabilistic models. Unlike us, the source of the knowledge in their informed neural process framework comes from an LLM-generated representation of the knowledge, (e.g. "This plot is increasing"), and they meta-learn the relationship between the knowledge embedding and the function space prior itself. Therefore their framework would not be applicable to our setting, as this form of knowledge integration would not help with validating existing theoretical models of neural systems.

\newpage

\section{Extended Figure 3a and Ablation Study Results}
\label{sec: ablations}
\begin{figure}[H]
    \centering
    \includegraphics[width=1\textwidth]{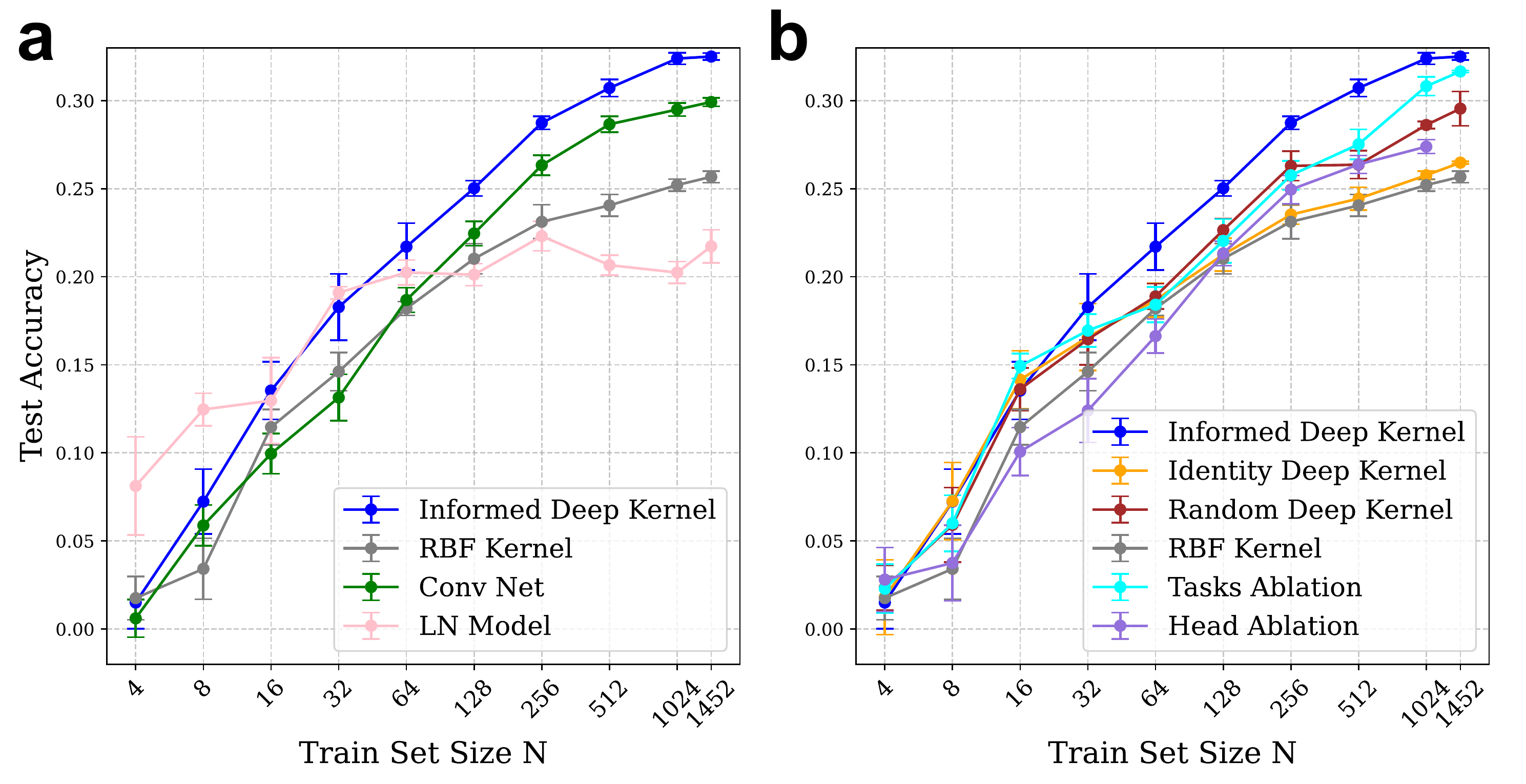}
    \caption{\textbf{Extended Figure 3a and Ablation study} (\textbf{a}) Reproduction of Figure 3a. (\textbf{b}) Accuracies of ablation models on biological data.}

    \label{fig:extendedresults}
\end{figure}

\begin{table}[h]
\centering
\small
\begin{tabular}{rccccc}
\toprule
$\!N$ & 

      Identity & RBF & Random & TaskAblation & HeadsAblation \\
\midrule
4     & 2.84e-01 & 6.56e-01 & 4.34e-01& 8.32e-01  & 8.85e-01\\
8     & 4.39e-01 & 1.89e-03*** & 2.67e-02** & 1.50e-01 & 4.50e-05***\\
16    & 8.02e-01 & 3.51e-02* & 3.37e-01 & 9.40e-01 & 5.97e-05***\\
32    & 1.85e-02* & 1.39e-03** & 2.12e-02* & 2.23e-02* & 2.77e-10***\\
64    & 8.29e-04*** & 6.14e-03** & 6.77e-03** & 1.37e-03** & 1.05e-07***\\
128   & 1.38e-04*** & 2.35e-03** & 5.04e-03** & 2.17e-03** & 3.19e-05***\\
256   & 6.26e-06*** & 6.84e-04*** & 5.99e-03** & 3.87e-03** & 2.06e-06***\\
512   & 4.43e-07*** & 8.75e-05*** & 4.74e-05*** & 2.48e-03** & 2.98e-07***\\
1024  & 6.67e-09*** & 1.87e-05*** & 3.37e-04*** & 1.10e-01 & 1.15e-07***\\
1452  & 7.16e-07*** & 9.37e-05*** & 5.54e-04*** & 2.47e-01 & - \\

\bottomrule

\end{tabular}
\caption{$p$-values for the one-sided paired Wilcoxon test (Hypothesis: Informed $>$ Control) on the 86 task-level means at each training-set size. 
\textsuperscript{*}\,$p<0.05$,\; 
\textsuperscript{**}\,$p<0.01$,\; 
\textsuperscript{***}\,$p<0.001$.}
\label{tab:wilcoxon}
\end{table}

\paragraph{Discussion} 

Figure \ref{fig:extendedresults} extends Figure 3a from the main text, now including the results from our ablation studies ('Identity', 'Random', 'Head Ablation', 'Tasks Ablation').

The LN model, a common baseline for modeling retinal responses, demonstrates strong performance in the extremely low-data regime. Notably, while the LN model is broadly superior in extremely small data regimes ($N<16$), our theory-informed model catches up in performance ($16\leq N\leq 64$), and subsequently exceeds that of the LN model - suggesting that the transferred theoretical knowledge provides an effective bias that helps in data-limited regimes but does not hinder as data becomes more abundant.

\paragraph{Ablation Studies}

To isolate the contribution of the theory-derived meta-learned feature extractor ($\phi$), we performed four ablation studies (Figure \ref{fig:extendedresults}b):
\begin{enumerate}
    \item An 'Identity' deep kernel ablation: This removes the feature extractor entirely, applying the task-adaptive linear heads directly on input pixels.\\
    
    \item A 'Random' deep kernel ablation: This randomizes the weights of $\phi$ (sampling from the standard initialization distribution), but keeps the architecture (Table \ref{tab:simplecnn}) fixed.
    
    This ablation represents a powerful comparison target, as large randomly initialized networks, particularly preceding learnable layers like our linear head and RBF GP, have been shown to represent features that capture relevant data structure, preserving input geometry, and in some cases nearly matching the performance of fully trained features ( \citep{rahimi2007random, rudi2017generalization, vershynin2018high,saxe2011random,ulyanov2018deep}).
    \item A "Task Ablation" condition on the meta-learning process, where we retrain the model, replacing the theory-derived set of meta-training tasks with a generic tasks derived from the statistics of natural images alone. 
    We replaced our meta-training set tasks with the tasks of trying to predict the values of the top principal components of natural images- with each task constituting one principal component. We kept all other conditions fixed. 
    \item A "Heads ablation" condition, where we remove the task adaptive linear heads
\end{enumerate}

\textbf{Statistical tests}\ \  We quantitatively compared the performance of our informed model against these two ablations (and the RBF GP) using a one-sided paired Wilcoxon signed-rank test \citep{wilcoxon1992individual} (hypothesis: Informed > Control). For each training-set size $N$, we first averaged the test correlation over the five random seeds within each task, yielding 86 paired observations for the test (one per task). 

The  $p$-values are presented in Table~\ref{tab:wilcoxon}, and statistical significance is annotated visually using asterisks in Figure ~\ref{fig:extendedresults}b. Statistical significance ($p < 0.05$) emerges for all baseline comparisons at $N \ge 32$, with high significance levels ($p < 10^{-3}$) consistently observed for training-set sizes of $N \ge 64$. For reference, the median correlation improvement over random features at $N = 512$ was $0.049$ (95\% BCa bootstrap confidence interval: $[0.018, 0.083]$), demonstrating that the detected differences are both statistically robust and practically meaningful.

The analysis confirms that our theory-informed model significantly outperforms the standard RBF GP baseline and the other ablation conditions across most training data sizes. The significant improvement over the 'Random' baseline is noteworthy; given that randomly initialized deep networks themselves provide strong feature representations (as evidenced by the strong performance of this baseline). The specific features obtained through meta-learning therefore confer a substantial structural advantage. Furthermore, the superior performance of both informed and random features compared to the 'Identity' baseline confirms the significant contribution of the deep feature extractor component itself, beyond the task-adaptive linear heads. Finally the superior performance of the theory-informed meta-learned features over theory-agnostic meta learned features (designed to encourage good generic features of natural images) shows that the accuracy benefit is specific to the theory-informed tasks. 

\textbf{Take Away:}\ \  Our ablation study supports the claim made in the main text: the meta-learned features effectively transfer knowledge from the theory, providing a beneficial inductive bias that leads to improved performance.

\subsection{Comparison to Transfer Learning}
We here highlight why we specifically need to use meta learning for our framework, as opposed to simpler mechanisms of knowledge transfer such as transfer learning. In standard Transfer Learning (TL), prior knowledge is encoded as a static feature set. This feature set is then re-used or fine-tuned by a simple downstream predictor on another task: directly constituting part of the prediction.

Our framework operates on a different principle entirely. While there IS a frozen feature extractor involved, what we transfer is a meta-learned theory-informed Bayesian prior over the space of functions. The role of the feature extractor is \textbf{not} to provide part of the solution for a task (e.g. by learning a center surround receptive field), but to specify an intricate function space prior out of an exceptionally vast set of functions representable by an abstract metric embedding.

Intuitively, transfer learning is like memorizing a part of the “answer” for one task, and transferring it onto another task. In contrast, our meta learning framework constructs, then transfers a “scheme” for how to find the answer for any task in a given domain.

This difference is consequential, conferring the following \textbf{advantages of our Meta Learning framework over Transfer learning}:
\begin{itemize}
    \item \textbf{Greater Generality}: As a prior over functions, our method is agnostic to exactly -what kind- of knowledge it needs to encode. While for fitting an RGC response to images, one could just as well invent clever regularizations on CNN filters to achieve similar or superior accuracy gains, this approach is limited to forms of knowledge that can be expressed as specific neural network weight values/statistics. In contrast, our prior can capture more abstract forms of knowledge, for example abstract rules for a rule-based decision making task. We discuss this more in Sec. \ref{sec: generality}
    \item \textbf{Task-Adaptive Prior} Unlike transfer learning, where the knowledge transfer is fixed, our theory informed prior can adaptively relax itself or absorb more informative structure via the linear heads. This enables generalization from idealized efficient coding predictions to the real, noisy biological data.
    \item \textbf{Principled Theory Validation and Uncertainty Quantification}  In our informed GP, we can compute the exact marginal likelihood of biological data under the theory-informed prior— a direct measure of the evidence for a given theory in units of information. This enables a direct quantitative comparison between competing scientific theories. Our GP is also probabilistic: presented with an image that its training data is not representative of, it outputs a high degree of uncertainty (Fig 3b,c), and not just a confidently wrong answer. 

\end{itemize}

\begin{figure}[h!]
    \centering
    \includegraphics[width=0.45\textwidth]{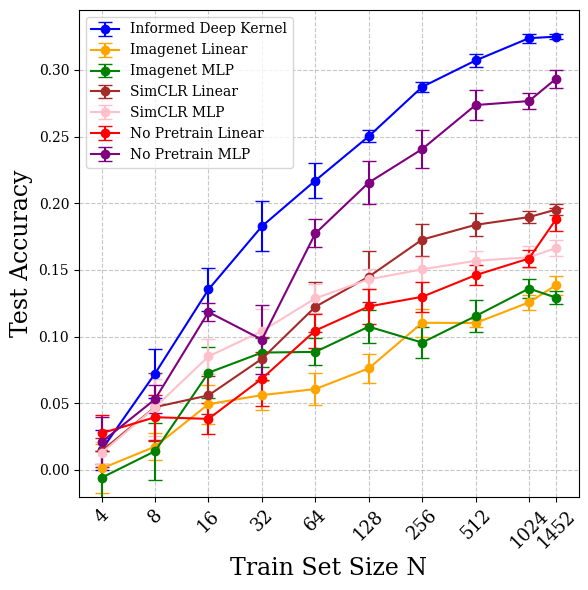}
    \caption{\textbf{Transfer Learning baselines} }

    \label{fig:TLbaselines}
\end{figure}

\paragraph{Transfer Learning Baselines}
We ran an extensive grid search to fine-tune linear and MLP heads on top of frozen features from pre-trained ResNet-18 models (using both ImageNet and SimCLR trained weights). As shown in our figure , we found that these pretrained features were not competitive, performing worse than even the plain MLP on the raw data (Figure \ref{fig:TLbaselines})

We hypothesize this is because the features from these deep networks are too abstract for RGC responses, a finding consistent with prior work ( \citep{cadena2019deep}) which found earlier-layer features were necessary even for modeling V1.

\section{On the Generality of our framework}
\label{sec: generality}

While our manuscript presents a single challenging deep-dive case study of the retina application, our framework is, by design, quite general.

\subsection{Deep Kernels define broadly applicable function space priors}
The generality of the Deep Kernel architecture stems from a core principle: \textbf{many complex biological phenomena are governed by a small number of simple, underlying latent variables.} Systems like these can be well described by our model.

To elaborate, metric-based methods confer significant flexibility when used to define function space priors in a deep kernel Gaussian process setting. 
Since the core of our model is a Gaussian Process with learned kernel $K_i(x,x') = \sigma^2_i\exp[ - \frac{||h_i\psi(x)-h_i(\psi(x'))||^2}{2l_i^2}]$, the prior over functions is fundamentally defined by the geometry of the embedding space learned by $\psi$ and refined by $h_i$. 

The core assumption of this architecture is that the deep neural network can learn a transformation of the potentially complex input space $\mathcal{X}$ to an embedding space $\mathcal{Z}$ where the functional relationships become simple, smooth, and stationary. The RBF kernel enforces a strong prior for smoothness in this learned space Z: if two points $h_i(\psi(x))$ and $h_i(\psi(x'))$ are close in Euclidean distance, their corresponding function values $f(x)$ and $f(x')$ are modeled to be highly correlated.

Therefore, the space of priors that can be fruitfully expressed is the set of all functions that can be simplified to a stationary smoothness assumption through a continuous mapping. This architecture is naturally suited for problems where the data generating function $f_{dg}(x): \mathbb{R}^D \rightarrow \mathbb{R}$ exhibits a specific structure.  The hypothesis is that while $f_{dg}$ may be complex in the high-dimensional input space $\mathcal{X}$, its structure is fundamentally governed by a smaller number of task-specific continuous latent variables, $v$. More formally, we assume there exists a low-dimensional manifold $\mathcal{M}$ parameterized by latent coordinates $v\in\mathcal{V}$ (where dim($\mathcal{V}$)$<$dim($\mathcal{X}$)) via a map $g:\mathcal{V}\rightarrow\mathcal{M}\subset\mathcal{X}$. The core hypothesis is that the true function can be expressed as the composition of a "simple" function $f_{simple}:\mathcal{Z}\rightarrow\mathcal{R}$ and the inverse of the parameterization, $g^{-1}$ : $f_{dg}(x) \approx f_{simple}(g^{-1}(x))$ for  $x\in \mathcal{M})$

The role of the deep network and linear head is to help approximate this $g^{-1}$, allowing \emph{our} $f_{simple}$ (an RBF Gaussian Process) to be a good prior for the data generating function. 

This is closely related the \textit{manifold hypothesis} in machine learning \citep{bengio2013representation,calandra2016manifold}, and is extraordinarily general across natural science datasets. \textbf{We provide numerous examples} in Table \ref{tab:generality table}.

\subsection{Generality via a Modular Meta-Learning Pipeline}

The generality of our work stems not only from the deep kernel architecture but from the underlying meta-learning pipeline itself. The framework's components are modular and can be adapted to problems of varying complexity and data modality.

\paragraph{Adapting the Feature Extractor to Data Modality:} While we employ a CNN for image data, the feature extractor $\psi$ is a pluggable module. For problems with different data structures, it could be replaced with a more suitable architecture, such as a Graph Neural Network for molecular data or a Transformer for sequence data, without changing the overall framework.

\paragraph{Adapting the Kernel to Problem Complexity:} The choice of a deep kernel is motivated by our complex, high-dimensional problem. For simpler scientific problems defined on low-dimensional inputs (e.g., modeling 1D orientation tuning curves), a deep feature extractor would be unnecessary. In such cases, our framework could be instantiated by meta-learning the hyperparameters of an expressive classical kernel, such as the \textbf{Spectral Mixture Kernel} \citep{wilson2013gaussian}, directly from theory-generated tasks. This demonstrates that the core principle—meta-learning a prior from a theory—is broadly applicable across different model complexities.
\newpage

\begin{table}[H]
\centering
\begin{tabularx}{\linewidth}{ l X X X X }
\toprule
\textbf{System to Model} & \textbf{Input $\mathcal{X}$} & \textbf{Output $y$} & \textbf{Latent Variable $\mathcal{V}$} & \textbf{(Normative) Theory} \\
\midrule
Primary Visual Cortex (V1) & Images & Single Neuron firing rates & Stimulus features (orientation, spatial frequency, motion) & Sparse Coding \citep{olshausen1996emergence} \\
Primary Motor Cortex (M1) & Neural Population Activity & Limb Velocity vector & Intended Limb Kinematics (Position, Velocity, Acceleration) & Optimal Feedback Control \citep{todorov2002optimal,vargas2024task} \\
Primary Auditory Cortex (A1) & A sound's time-frequency spectrogram & Single Neuron firing rate & Spectro-temporal features like pitch, timbre & Efficient Coding of Natural Sounds \citep{lewicki2002efficient} \\
Hippocampal Place Cells & Multi-Sensory inputs & Place Cell firing & The animal's 2D spatial position, proximity to abstract goals \citep{el2024cellular}, and head direction. & Successor Representations \citep{momennejad2017successor} \\
Prefrontal Cortex (PFC) & Stimuli and Reward history & PFC population activity & Abstract Task Variables ("belief state", "decision rule") & Bayesian Decision theory \citep{kording2006bayesian}, Drift Diffusion models \\
Protein Science & Amino Acid Sequence & A functional property (e.g., catalytic efficiency, stability) & Geometric or electrostatic properties of the protein's active site & Molecular Dynamics simulations \\
Molecular Property prediction & Small Molecule features (Connectivity Fingerprint) & reactivity (EC50) & pharmacophores or structural motifs & DFT or QSAR predictions\\
Developmental Biology & Vector of local morphogen concentrations & Cell-fate gene expression level & Cell's physical position along a developmental axis & Reaction-Diffusion Models \citep{zagorski2017decoding} \\

\bottomrule
\end{tabularx}
\caption{A selection of scientific systems where the core hypothesis of our framework may be applicable.}
\label{tab:generality table}
\end{table}

\section{Biological Data Preprocessing}
\label{sec: biological_data}
\subsection{Dataset Source and Initial Processing by Qiu et al.}
Our dataset was sourced from the published work of Qiu et al \citep{qiu2023efficient,qiu2023dataset}, focusing specifically on the responses of 86 mouse ventral ganglion cells to a 30Hz natural movie stimulus with Green and Ultraviolet (UV) channels, as described in our main text. To reiterate, each input to our regression model is derived from the downsampled image frames (36$\times$32 pixels) presented at 30Hz to the retinal cells. The outputs we predicted are the Calcium fluorescence responses measured for each neuron, acquired via two-photon microscopy at a frame rate of 7.8125 Hz. These fluorescence signals serve as proxies for neuronal firing activity, as calcium indicators reflect intracellular calcium concentration changes associated with cells firing action potentials.  While we applied our own our processing steps (temporal flattening via a canonical filter derived from an LN model, hierarchical clustering to select unique representative images, selection of the UV channel, and z-scoring), Qiu et al. performed crucial initial preprocessing on the data, most important of which we describe here- but refer readers to their publication for full details.

\paragraph{Cell Selection via Quality Index (QI)}
The dataset includes responses from cells selected based on their response reliability. Qiu et al. calculated a Quality Index (QI), ranging from 0 to 1, for each cell based on the consistency of its responses across repeated presentations of specific test sequences. The QI was calculated as:
$$ QI = \frac{\text{Var}_tx \left( E_r [C] \right)}{E_r \left( \text{Var}_t [C] \right)} \quad $$
where $C$ represents the response matrix with dimensions time samples ($t$) by repetitions ($r$), $E_r[\cdot]$ denotes the expectation (average) across repetitions, and $\text{Var}_t(\cdot)$ denotes the variance across time samples. A higher QI indicates a more reliable response with a higher signal-to-noise ratio. For inclusion in the final dataset used for modeling, neurons responding to the natural movie stimulus had to meet the criterion $QI_{\text{noise}} > 0.25$, where $QI_{\text{noise}}$ was determined using responses to separate noise stimulus test sequences also presented during the experiments.

\paragraph{Stimulus and Response Handling}
Importantly for evaluation, the 750 test and validation fluorescence responses provided in the dataset are not raw signals but represent an average across 3 repetitions of specific held-out input sequences from the natural movie stimulus. This averaging serves to increase the signal-to-noise ratio of these benchmark traces. This makes the original train/test+val split non-interchangeable.

For exhaustive details on the experimental procedures (including animal handling, tissue preparation, recording setup) and the complete preprocessing pipelines applied across all stimulus conditions in the original study, we refer the reader to the source publications \citep{qiu2023efficient,qiu2023dataset}. Our subsequent processing steps applied for this work are detailed below.

\subsection{Temporal Filter Extraction via LN Model}
\label{sec: biological_LN_filter}

\begin{figure}[h!]
    \centering
    \includegraphics[width=0.8\textwidth]{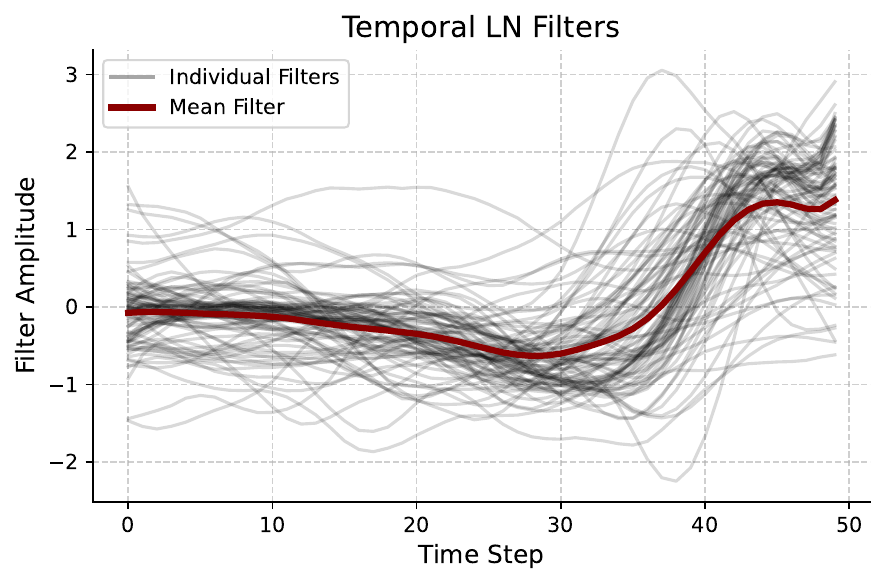}
    \caption{\textbf{Time Filters of LN model on Biological Data} Individual filters shown are post-flipping. }

    \label{fig:LNtimefilters}
\end{figure}

To convert the regression problem from movie inputs to image inputs, we temporally flatten the movies. To achieve this, we first fit a spatiotemporal Linear-Nonlinear (LN) model to predict the 16200 neural responses $\in \mathbb{R}^{16200\times86}$ from 16200 movie inputs $\in \mathbb{R}^{16200\times1152\times50}$. This LN model consists of spatial filter parameters $W \in \mathbb{R}^{86\times1152}$ and Temporal filter parameters $\tau \in \mathbb{R}^{86\times50}$. Among temporal filters we observed classic biphasic shapes (dip-then-peak or vice versa). We then "flipped" some filters by multiplying by -1 in order to align their biphasic structures and allow for averaging, and then averaged across neurons to obtain a single "canonical" filter $F \in \mathbb{R}^{50}$, spanning 50 frames. The choice of which filters needed flipping was made by thresholding at the peak value. Each movie frame $x_\tau \in \mathbb{R}^{36\times32}$ was then projected onto this canonical temporal filter, effectively collapsing the time dimension into a single static ``image'', $\tilde{x} = \sum_{\tau=1}^{50} F_\tau x_\tau$. We plot all 86 time filters identified this way (Fig. \ref{fig:LNtimefilters}) alongside the canonical average filter.

\subsection{Filtering non redundant images}
\label{sec: biological_removing_duplicate}

To derive a representative subset from the 16,200 images in our dataset, we applied a data-driven hierarchical clustering approach. We first computed 1-NN distances (excluding self-matches) in the feature space (shaped 16200 × 1152) using Scikit-Learn's NearestNeighbors with Euclidean distance. These distances were sorted, and a clustering threshold was selected based on the largest gap between successive values—a heuristic for natural cluster separation. We then performed agglomerative clustering with average linkage and this threshold (with n$\_$clusters=None), resulting in 1,452 clusters. A single exemplar (the first image) from each cluster formed our representative subset.

\section{Synthetic Meta-Train Dataset Generation}
\label{app:synthetic_data}
This section outlines the generation of the synthetic dataset based on our chosen efficient coding theory.
\subsection{Normative Model (Efficient Coding Autoencoder)}
\paragraph{Architecture of Efficient Coding Autoencoder} Table \ref{tab:normativeautoencoder} shows the full architecture used.

\begin{table}[H]
\centering
\begin{tabular}{|c|c|c|c|}
\hline
\textbf{Layer} & \textbf{Details} & \textbf{Output Shape} & \textbf{Activation} \\
\hline
Input & - & B, 1, 28, 28 & - \\
\hline
Conv & 9×9, padding=4 & B, 16, 28, 28 & ReLU \\
\hline
Flatten & - & B, 12544 & - \\
\hline
Linear1 & 12544 → 300 & B, 300 & - \\
\hline
Gaussian Noise & $\sigma=2.0$ (train only) & B, 300 & - \\
\hline
Bottleneck & - & B, 300 & ReLU \\
\hline
Linear2 & 300 → 12544 & B, 12544 & ReLU \\
\hline
Unflatten & - & B, 16, 28, 28 & - \\
\hline
ConvTranspose & 9×9, padding=4 & B, 1, 28, 28 & Tanh \\
\hline
Output & - & B, 1, 28, 28 & - \\
\hline
\end{tabular}
\caption{\textbf{Normative Autoencoder Architecture} }
\label{tab:normativeautoencoder}

\end{table}

\paragraph{Training Procedure}

The autoencoder model, adapted from work by Ocko et al. and Qiu et al. \citep{ocko2018emergence,qiu2023efficient}, was trained to learn an efficient code for natural images. The core idea of efficient coding is that sensory systems aim to represent naturalistic stimuli faithfully while minimizing metabolic costs. Our autoencoder operationalizes this by optimizing the loss function $\mathcal{L}_{\text{EC}}$; defined as:

\begin{equation} \label{eq:autoencoder_loss_integrated}
\mathcal{L}_{\text{EC}} = \text{MSE}(X, \hat{X}) + \beta ||A_{\text{bottleneck}}||_1 + \alpha \left( ||W_{\text{conv}}||_2^2 + ||W_{\text{tconv}}||_2^2 \right)
\end{equation}
where $\text{MSE}(X, \hat{X}) = \frac{1}{N} \sum_{j=1}^{N} ||X_j - \hat{X}_j||_2^2$ is the mean squared error ensuring fidelity between original ($X_j$) and reconstructed ($\hat{X}_j$) images over a batch of $N=200$ samples. The second term applies an L1 penalty to the bottleneck layer activations $A_{\text{bottleneck}}$ with weight $\beta=1 \times 10^{-5}$ to promote sparse codes. The third term imposes an L2 penalty on the weights of convolutional ($W_{\text{conv}}$) and transposed convolutional ($W_{\text{tconv}}$) filters with weight $\alpha=5 \times 10^{-5}$ to encourage smooth filters. Optimization was performed over 200 epochs using the Adam optimizer (learning rate $1 \times 10^{-5}$). We trained on $60000$ patches sampled from the public Van-Hateren dataset of natural images \citep{van1998independent}.

Consistent with efficient coding frameworks that account for noise, additive Gaussian noise (std 2) was applied to the input images during training, simulating sensory noise and acting as a regularizer. While Ocko et al. \citep{ocko2018emergence} primarily used noise and sparsity penalties without an explicit narrow bottleneck in their model, our implementation imposed a fixed bottleneck size. To ensure we were meta-training on a high-fidelity representation of the theory, we \textbf{performed a small hyperparameter search} over the normative model's architecture and selected a 300-unit bottleneck, along with the other hyperparameter values. This was a pragmatic choice that reliably produced the canonical center-surround receptive fields predicted by efficient coding. Our chosen hyperparameters ($\alpha$, $\beta$, input noise, bottleneck size), represent one point in a large trade-off space between reconstruction accuracy, code sparsity, and filter smoothness, and were chosen to ensure a robust and faithful instantiation of the theory for our subsequent meta-learning. Varying these parameters would likely yield different efficient coding solutions, resulting in different filter properties (e.g. center/surround relative and absolute size) as the model adapts to different balances of these competing objectives under varying architectural and noise constraints.

\paragraph{Example Receptive Fields}

\begin{figure}[h!!]
    \centering
    \includegraphics[width=0.5\textwidth]{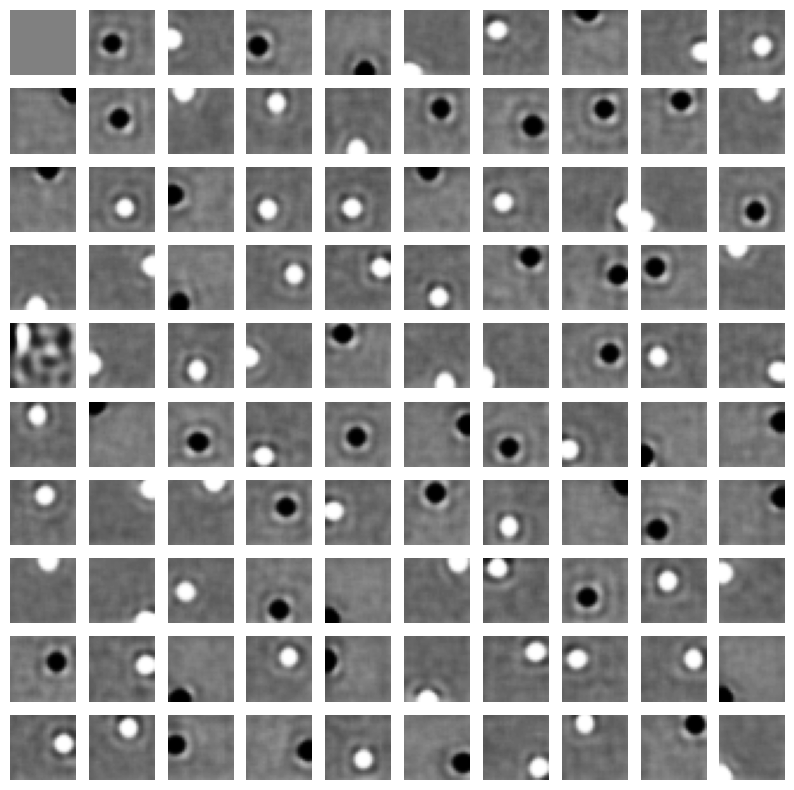}
    \caption{\textbf{Normative Autoencoder Receptive Fields}}

    \label{fig:normativerfs}
\end{figure}

To validate the features learned by the hidden units in the autoencoder's bottleneck layer, we visualized the preferred input pattern associated with each unit. This was achieved by activating each of the 300 bottleneck units individually (setting its value to 1 while others were 0) and passing this sparse representation through the trained decoder network. The resulting decoded output image represents the spatial pattern that maximally drives (or is reconstructed by) that specific bottleneck unit (Fig. \ref{fig:normativerfs})

\subsection{Data Augmentations Performed to extend the Meta-Train set}
To generate the synthetic "tasks", we extract spatial receptive fields of the "optimal" bottleneck neurons of our efficient coding autoencoder by fitting a spatial Linear-Nonlinear (LN) model to the activations of the autoencoder bottleneck. This yields 300 center-surround filters ${f_i}$ . Each filter then defines one synthetic task obtained by taking its dot product with the image stimuli $y_{i,k} = f_i^T x_k$ where $x_k$ is the $k'th$ image. 
We expand this meta-train set by choosing 20 archetypal un-obfuscated center surround filters (by visual inspection), \textbf{rescaling their size} (randomly ranging from 0.8x to 1.2x) and \textbf{jittering} the location of their 'centers' , allowing us to increase the size of the synthetic meta-train set $N_\mathcal{T}$. We created 4000 such synthetic tasks, but ultimately used only $N_\mathcal{T}=490$ selected randomly, as we found empirically that this number of tasks saturated the meta-training performance. 
\paragraph{Rationale behind Jittering and Scaling.} Our Jittering and Scaling augmentations are not arbitrary, but rather a principled computationally efficient way to generate new tasks consistent with the same normative theory. 
Our meta learning framework treats each neuron as an individual "task", meaning our framework learns a prior for \textbf{single-neuron} responses. Our meta-train set is therefore composed of many separate individual examples of "good" single neuron solutions consistent with the theory which is itself posed at a population level. On a single neuron level, efficient coding theories predict a \textit{type} of solution (a center-surround structure), not one single filter at one specific pixel location. Therefore, \textbf{a filter jittered to a different location is an equally valid example of a high-performing solution under the theory of Ocko et al}: it would have effectively the same utility level. This is in stark contrast to population level efficient coding theories, which may precisely specify the relative positions, overlap levels, and tilings of a population of efficient coding neurons- so if we were learning a population-level prior rather than a single neuron level prior, the jittering would be problematic as it would be deleting that information.

The \textbf{spatial scale of receptive fields}, on the other hand, is a known consequence of architectural capacity and regularization in efficient coding models. Ocko et al, for instance, have analytically connected the size of the emergent receptive fields to the "stride length" of the convolutional filters- driven by the need to prevent aliasing. Other seminal works, such as that of Karklin and Simoncelli \citep{karklin2011efficient}, have demonstrated that the "optimal" efficient receptive field size is a function of the input noise level. Because the noise is applied i.i.d to each pixel, a high level of noise primarily affects the Signal-to-noise ratio of high-frequency components of the image's power spectrum. Therefore, in a high-noise environment, the optimal solution is to average over a larger spatial area to filter out the noise. This corresponds to a larger receptive field. On the other hand, given low noise, the high frequency features are more reliable, and therefore valuable to capture with spatially smaller receptive fields.

There is not a single efficient coding theory predicting a single "optimally sized" canonical receptive field. The \textbf{optimal size is conditional} on the stimulus statistics and architectural constraints. In fact, biological retinal ganglion cells are known to adapt the sizes of their functional receptive fields via inhibitory plasticity precisely in response to these changing environmental statistics \citep{smirnakis1997adaptation}. Therefore, our choice to spatially scale the sizes of receptive fields in a reasonable range is a prinicipled, computationally efficient way to convert our Efficient Coding solutions trained at a single fixed noise level and architecture into a family of solutions consistent with a larger range of these constraints- which are in turn \underline{all} consistent with the theory- being optimal solutions corresponding to different hyperparameter values.

If we had known the precise noise levels experienced by the photoreceptor cells of the \textit{ex vivo} biological neurons, we may try to reverse engineer a single noise level in our model consistent with their true underlying noise level and test the predictions of the theory at that level of specificity. However, since that is far beyond our scope, \textbf{scaling the receptive fields the most principled thing to do}: resulting in a set of meta-train tasks representative of a wide range of these hyperparameter values which we as scientists do not know the ground truth value of.

Nonetheless, we provide an \textbf{Ablation study on the Scaling data augmentation}, demonstrating that our model's accuracy is, if anything, reduced by the preprocessing across most data sizes.

\begin{tabular}{r||c|c|c|c|c }

    N & 4 & 8 & 16 & 32 & 64 \\
    \hline
    
    Theory Informed Kernel &  $0.015 \pm 0.015$ & $0.072 \pm 0.019$ & $0.135\pm 0.017$ & $0.182\pm 0.019$ & $0.217\pm 0.014$ \\

    Ablation: No Scaling & $0.021 \pm 0.015$ & $0.079 \pm 0.022$ & $0.154 \pm 0.020$ & $0.192 \pm 0.022$ & $0.231 \pm 0.010$ \\
    
\end{tabular}

\begin{tabular}{r||c|c|c|c|c }

    N &   128 & 256 & 512 & 1024 & 1422\\
    \hline
    
    Theory Informed Kernel &  $0.250 \pm 0.005$ & $0.287\pm 0.004$ & $0.307\pm 0.005$ & $0.323\pm 0.004$ & $0.325 \pm 0.002$\\

    Ablation: No Scaling &  $0.256 \pm 0.008$ & $0.289 \pm 0.007$ & $0.305 \pm 0.005$ & $0.314 \pm 0.002$ & $0.324 \pm 0.001$\\
    
\end{tabular}

\section{Meta-Learning Algorithm and Training Details}
\label{app:meta_learning}

\subsection{Model Architecture}

\paragraph{Feature Extractor Architecture}
Table \ref{tab:simplecnn} shows our feature extractor architecture.

\begin{table}[H]
\centering

\begin{tabular}{|c|c|c|c|}
\hline
\textbf{Layer} & \textbf{Details} & \textbf{Output Shape} & \textbf{Activation} \\
\hline
Input & - & B, 1, 36, 32 & - \\
\hline
Conv1 & 3×3, padding=1 & B, 32, 36, 32 & GeLU \\
\hline
Conv2 & 3×3, padding=1 & B, 64, 36, 32 & GeLU \\
\hline
MaxPool & 2×2 & B, 64, 18, 16 & - \\
\hline
Conv3 & 3×3, padding=1 & B, 128, 18, 16 & GeLU \\
\hline
Conv4 & 3×3, padding=1 & B, 128, 18, 16 & GeLU \\
\hline
Flatten & - & B, 36864 & - \\
\hline
Linear1 & 36864 → 256 & B, 256 & GeLU \\
\hline
Linear2 & 256 → 256 & B, 256 & - \\
\hline
\hline
\end{tabular}
\caption{\textbf{CNN feature extractor ($\phi$) architecture}}
\label{tab:simplecnn}

\end{table}

\paragraph{Linear Heads}
All linear heads are all torch.nn.Linear instances, with 256 input dims and 128 output dims and L1 regularization (=0.01) to encourage generalization \citep{fumero2023leveraging}, also following the recommendation from \citep{van2021feature} to not overly shrink the latent dimensionalities.

\paragraph{Gaussian Process Layer}

For all of our GP layers, we used a Gpytorch Exact GP with an RBF Kernel  $k(x,x';\theta_{gp})=\sigma_{f}exp(-\frac{\|x-x'\|^{2}}{2l^{2}})$.
We set all mean functions to ZeroMean(), and used a Gaussian Lengthscale Prior with a variance of $0.01$ centered around the initialization lengthscale value.
For initialization, we followed the heuristic "median lengthscale init" reported in previous work (e.g. \citep{van2021feature}) where the lengthscale is initialized to be the median distance between the embedded data points. Importantly, we did \emph{not} keep re-computing the median init as we cycled through task batches, instead saving the first computed median and re-initializing with that value. Experimental validation demonstrated the importance of this fixed median initialization and narrow lengthscale hyper-prior for avoiding feature collapse (Sec. \ref{featurecollapse})
\subsection{Meta-Training Procedure}

\begin{algorithm}
\caption{Procedure for Fitting Adaptive Deep Kernel Model}
\label{alg:adaptive-deep-kernel}
\begin{algorithmic}[1]
  \scriptsize
  \STATE \textbf{Define:} Task-adaptive parameters $\theta_{\text{adapt}}$ (i.e. $\theta_{\text{gp},i}$ and  $W_i$ from $h_i$) 
  \STATE \textbf{Define:} Meta-learned parameters $\theta_{\text{meta}}$ (i.e. $\theta_{\text{nn}}$ from $\phi$)
  \STATE \textbf{Input:} Meta training set 
  \[
    \mathcal{T} = \Bigl\{ \mathcal{T}_i = \Bigl( \mathcal{D}_{i}^{\text{support}},\, \mathcal{D}_{i}^{\text{query}} \Bigr) \Bigr\}_{i=1}^{N_{T}},
  \]
  where for each task $\mathcal{T}_i$:
  \[
    \mathcal{D}_{i}^{\text{support}} = \{(x^{\text{support}}_{i,k},\, y^{\text{support}}_{i,k})\}_{k=1}^{n_i^{\text{support}}} \quad \text{and} \quad \mathcal{D}_{i}^{\text{query}} = \{(x^{\text{query}}_{i,k},\, y^{\text{query}}_{i,k})\}_{k=1}^{n_i^{\text{query}}}.
  \]
  \STATE Shuffle $\mathcal{T}$ and partition into batches
  \FOR{epoch $=1$ \TO $E$}  
    \FOR{batch $\in$ $\mathcal{T}$}
          \FOR{each task $\mathcal{T}_i = \bigl(\mathcal{D}_{i}^{\text{support}},\, \mathcal{D}_{i}^{\text{query}}\bigr)$ in the batch}
            \STATE Reinitialize task-adaptive parameters $\theta_{\text{adapt,i}}$

            \FOR{$t = 1$ \TO $n_{\text{inner steps}}$}
              \STATE Update task-specific parameters:
              \[
                \theta_{\text{adapt}, i} \gets \theta_{\text{adapt}, i} - \alpha_{\text{inner}} \, \nabla_{\theta_{\text{adapt}}} \log P\Bigl(y^{\text{support}}_i \mid x^{\text{support}}_i, \theta_{\text{adapt}, i} ,\theta_{\text{meta}} \Bigr ),
              \]
              where $\alpha_{\text{inner}} = (\alpha_{\text{linear}},\, \alpha_{\text{GP}})$ denotes the learning rates for the linear head and GP hyperparameters, respectively.
            \ENDFOR
            \STATE Set $\theta_{\text{adapt}, i}^* \gets \theta_{\text{adapt}, i}$.
        
          \ENDFOR
          \FOR{$j = 1$ \TO $n_{\text{outer steps}}$}
            \STATE Update the meta-parameters:
            \[
              \theta_{\text{meta}} \gets \theta_{\text{meta}} - \alpha_{\text{outer}} \, \nabla_{\theta_{\text{meta}}} \mathbb{E}_{i \in \text{batch}} \log P\Bigl(y^{\text{query}}_i \mid x^{\text{query}}_i, \mathcal{D}_{i}^{\text{support}}, \theta_{\text{adapt}, i}^*, \theta_{\text{meta}}\Bigr),
            \]
            where $\alpha_{\text{outer}}$ is the meta-learning rate.
        \ENDFOR
        \ENDFOR
        \ENDFOR

\end{algorithmic}
\end{algorithm}

\vspace{1em} 

\begin{table}[H]
\centering
\caption{Hyperparameters for the Optimization Procedure (Algorithm~\ref{alg:adaptive-deep-kernel})}
\label{tab:hyperparameters}
\begin{tabular}{llll}
\toprule
\textbf{Hyperparameter} & \textbf{Symbol} & \textbf{Description} & \textbf{Value} \\
\midrule
Number of Epochs & $E$ & Total number of meta-training epochs. & 7 \\
Batch Size & $B$ & Number of tasks per meta-batch. & 10 \\
Inner Adaptation Steps & $n_{\text{inner steps}}$ & Number of gradient steps for task adaptation. & 17 \\
Inner Learning Rate (Linear) & $\alpha_{\text{linear}}$ & Learning rate for linear head parameters in inner loop. & 0.0004 \\
Inner Learning Rate (GP) & $\alpha_{\text{GP}}$ & Learning rate for GP hyperparameters in inner loop. & 0.002 \\
Outer Meta-Update Steps & $n_{\text{outer steps}}$ & Number of outer loop optimization gradient steps  & 3 \\
Outer Learning Rate & $\alpha_{\text{outer}}$ & Learning rate for meta-parameters ($\theta_{\text{meta}}$) in outer loop. & 0.0004\\
\bottomrule
\end{tabular}
\end{table}

Our training procedure (Alg. \ref{alg:adaptive-deep-kernel}) is similar to that of the "Adaptive Deep Kernel Fitting via the Implicit Function Theorem" (ADKF-IFT) method by Chen et al \citep{chen2023meta}. We employ Bi-level optimization: with disjoint task-adaptive and meta-learned modules trained separately, on different learning objectives. Like Chen et al, we use the standard GP marginal log likelihood (Alg. \ref{alg:adaptive-deep-kernel},Eq. 1, main paper) for our "inner" training loop on task-adaptive parameters and the log-probability of the query data for the "outer" training loop on meta-learned parameters (to encourage generalization \citep{lotfi2022bayesian}). Unlike Chen et al, we do not use implicit differentiation to enrich the outer level gradient, as this would not be computationally feasible with our significantly higher number of task-adaptive parameters. Future work may explore the likely benefits of employing a scheme to approximate these implicit gradient terms \citep{geng2021training}. We also make adjustments to the training process to avoid feature collapse, discussed in \ref{featurecollapse}

\begin{figure}[h!]
    \centering
    \includegraphics[width=1\textwidth]{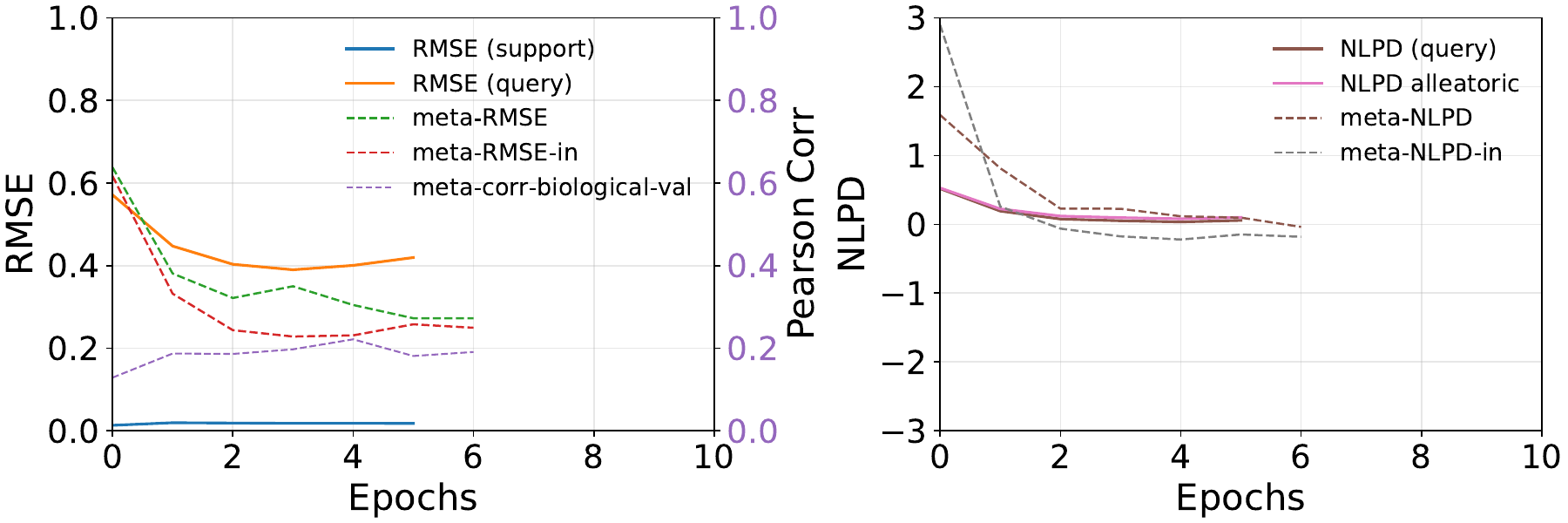}
    \caption{\textbf{Training Stats},Left Plot, Left Axis corresponds to mean RMSE (averaged over tasks), right axis corresponds to pearson correlation averaged over tasks (biological). Right plot depicts NLPD values, with NLPD(query) depicting epistemic NLPD and NLPD alleatoric including the noise term (which here is set to be very low). Dotted lines signify post-adaptation tests of the learned feature extractor at the start of each epoch. Solid lines are stats accumulated during training. The dotted lines are evaluated at the start of every epoch, while the solid lines are evaluated at the end of every epoch, which is why there is one extra dotted line point. Meta RMSE and Meta NLPD correspond to unseen tasks drawn from the same distribution, and meta-rmse-in and meta nlpd in correspond to tasks from the meta train set. To be clear, we early-stopped using the dotted purple line, which reflects what a scientist using our method should/would do.}

    \label{fig:trainingstats}
\end{figure}

We trained on batches of 10 (synthetic) tasks at a time, each task containing 1452 labeled input/output training data, randomly sampling 49 such batches, resulting in a total of 490 tasks in our meta train set. 
We trained for 7 epochs in total. At the start of every epoch, we ran the full adaptation procedure on three meta sets, using 100 support points each. One on Synthetic data from the meta train set. One on Synthetic data unseen during meta training but from the same distribution. One on our Biological data, with all evaluations computed on the validation data only, using a subset of tasks. We employed early stopping based on held-out biological data Pearson correlation (Fig. \ref{fig:trainingstats}). During our training process, we clipped the gradients on the base feature extractor to have a maximum norm of 1, and we trained the first epoch with 1/10th the learning rate. As shown in Alg. \ref{alg:adaptive-deep-kernel}, we optimized the feature extractor parameters $\theta_{nn}$ on the log-probability of the query data (outer level) in the meta-train set and the linear heads / GP layer parameters ($W_i$ and $\theta_{\text{gp},i}$) on the MLL of the support set (inner level). For synthetic meta train sets, we manually set the GP noise level close to zero to avoid feature collapse- preventing the log-prob objective from hiding collapsing epistemic uncertainty behind high alleatoric uncertainty (as discussed in \ref{featurecollapse}).
Our experiments indicated that utilizing a smaller proportion of data points for the support set, with a correspondingly larger proportion for the query set, fostered more effective representation learning in the feature extractor. In our case, out of 1452 possible points to be divided into both for each task, we used $5\%$ of them for our support set, and $95\%$ for our Query set, forcing the model to predict more unseen points using fewer conditioned values, and re-sampling the exact support/query split every epoch. We also adhered to the torch.double() dtype to avoid numerical errors from the low noise level.

\subsection{Task Adaptation Procedure} 
\label{sec:taskadaptationprocedure}
For each meta-test task (i.e., each biological neuron), we perform task adaptation to fit the model to the available biological data (the support set for that task). This involves optimizing only the task-specific parameters while keeping the meta-learned feature extractor weights frozen. The parameters adapted for each task $i$ include the weights of its linear head ($W_i$) and the hyperparameters of its Gaussian Process (GP) layer $\theta_{gp,i} = (\sigma_{f,i}, \ell_i, \sigma_{\eta,i})$, encompassing the GP output scale, length scale, and likelihood noise variance, respectively.

Adaptation is performed by maximizing the marginal likelihood of the support data for that task using the Adam optimizer \citep{kingma2014adam} for 300 epochs. We used a base learning rate of $4 \times 10^{-3}$ for the GP hyperparameters and Adam betas of $(0.99, 0.999)$. The learning rate for the linear head weights $W_i$ was scaled down by a factor of $10^{-2}$ relative to the base rate (i.e., $4 \times 10^{-5}$). Additionally, an L1 sparsity penalty with a coefficient of $10^{-2}$ was applied to the linear head weights during adaptation. For the GP hyperparameters, the length-scale prior was kept narrow (Gaussian with variance 0.01, except for the RBF-only case) and its mean was initialized using the median heuristic applied to the pairwise distances of the features $d(x,x') = ||h_i(\phi(x))-h_i(\phi(x'))||_2$ from the frozen extractor and freshly initialized heads. For experiments involving synthetic meta-test data, the GP likelihood noise $\sigma_{\eta,i}$ was initialized at $10^{-4}$- reflecting known ground truth. For biological data, the standard GPyTorch initialization was used.

Crucially, these same adaptation settings (optimizer, learning rates, epochs, priors, initializations) were used consistently across our main 'Informed' model and all ablation conditions. Given the significant computational cost required for extensive hyperparameter tuning for each model variant across dozens of tasks and multiple data subsets, we adopted these fixed, reasonable settings based on preliminary experiments. The consistent application ensures a fair comparison between the conditions, and the sensible trends observed in the results (e.g., Fig. \ref{fig:extendedresults}) suggest that these settings, while perhaps not perfectly optimal for every single task/model pairing, were adequate and did not unduly handicap any specific model.

\subsection{Design Choices Mitigating Feature Collapse in our Deep Kernel model}
\label{featurecollapse}

Although literature such as \citep{chen2023meta} suggests that meta-learning across diverse tasks might inherently mitigate feature collapse, our experiments indicated this was not guaranteed for our use case. However, we successfully avoided feature collapse in our model, attributing this primarily to three specific design choices in our training methodology:

\begin{enumerate}
    \item \textbf{Narrow Hyperprior on GP Lengthscales:} We imposed a tight prior distribution on the Gaussian Process (GP) lengthscale parameters ($\ell_i$), severely restricting their ability to adapt during the inner-loop optimization for each task.
    \item \textbf{Fixed GP Lengthscale Initialization:} We did not re-initialize the GP lengthscales using the median heuristic at the start of each task batch adaptation, a technique commonly used to set reasonable initial values \citep[e.g.,][]{chen2023meta,van2021feature}. We used the median heuristic to compute the lengthscale once, at the start of the first task batch, and then used that lengthscale throughout the other taskbatches.
    \item \textbf{Epistemic Log Probability Objective:} Our outer-loop meta-training objective optimized the feature extractor ($\phi$) based on the log probability of the query data ($y_q$) conditioned on the support data ($(x_s, y_s)$), often referred to as the negative log predictive density (NLPD). Unlike existing training approaches, however, we explicitly used the log probability of the data excluding the likelihood noise.
    This focuses purely on the epistemic uncertainty captured by the GP, unlike objectives combining epistemic and aleatoric uncertainty \citep[e.g.,][]{chen2023meta,lotfi2022bayesian} or optimizing the log marginal likelihood (LML) directly \citep[e.g.,][]{patacchiola2020bayesian}.
\end{enumerate}

Our preliminary understanding of the mechanisms by which these choices prevent collapse follows:

\textbf{Choices 1 and 2 (Lengthscale Restriction):} We observed empirically that allowing the GP lengthscale ($l_i$) to be freely optimized within each task's inner loop (as is typical in GP regression and done for task-adaptation in \citep{chen2023meta}) or frequently re-initializing it via the median heuristic, created a detrimental positive feedback loop during meta-training. Shorter lengthscales appeared to incentivize the deep feature extractor ($\phi$) to map inputs closer together in feature space. This proximity, in turn, further encouraged shorter lengthscales via the inner-loop LML optimization, ultimately leading to collapsed feature representations where distinct inputs map to nearly identical points. By fixing the lengthscale via a narrow prior and consistent initialization, we broke this feedback loop. The potential loss of flexibility from a fixed lengthscale was compensated by the adaptability of the meta-learned feature extractor ($\phi$) and the task-specific linear heads ($h_i$), which could still arrange the features appropriately for the given lengthscale.

\textbf{Choice 3 (Epistemic Log Probability Objective):} This choice appears to counteract feature collapse through two mechanisms:

\textbf{Avoiding Prior Log-Determinant Minimization:} Ober et al. \citep{ober2021promises} identified a key mechanism driving feature collapse in DKL models trained via LML maximization. The LML objective can be decomposed into a data fit term and a complexity penalty term:
    $$ \log p(y) = \underbrace{-\frac{1}{2}y^{T}(K+\sigma_{n}^{2}I_{N})^{-1}y}_{\text{Data Fit Term}} \underbrace{- \frac{1}{2}\log|K+\sigma_{n}^{2}I_{N}|}_{\text{Complexity Penalty Term}} - \frac{N}{2}\log(2\pi) $$

    Crucially, Ober et al. prove (in their Proposition 1) that for kernels with a learnable output scale ($\sigma_f$), the data fit term becomes a constant ($-N/2$) at the optimum. Consequently, maximizing the LML with highly flexible feature extractors becomes dominated by minimizing the complexity penalty, specifically the log determinant term $\log|K+\sigma_{n}^{2}I_{N}|$. The model achieves this by learning feature mappings ($\phi$) that make the kernel matrix $K$ (evaluated on the learned features) as close to singular as possible, which corresponds to collapsing the features.

    Our objective, however, focuses on the log probability of query data $y_q$ given support data $(X_s, y_s)$, which depends on the \emph{posterior} predictive distribution for the latent function values $f_q$ at query inputs $X_q$. This distribution is Gaussian, $p(f_q | X_q, X_s, y_s) = \mathcal{N}(f_q | \mu_{q|s}, \Sigma_{q|s})$. Let $X_s = \{x_{s,1}, \dots, x_{s,n_s}\}$ be the set of $n_s$ support inputs and $X_q = \{x_{q,1}, \dots, x_{q,n_q}\}$ be the set of $n_q$ query inputs. The kernel function, parameterized by the feature extractor $\phi$, is $k_{\phi}(\cdot, \cdot)$. We define the following kernel matrices:
    \begin{itemize}
        \item $K_{ss} = [k_{\phi}(x_{s,i}, x_{s,j})]_{i,j=1}^{n_s}$, the $n_s \times n_s$ kernel matrix between support points.
        \item $K_{qq} = [k_{\phi}(x_{q,i}, x_{q,j})]_{i,j=1}^{n_q}$, the $n_q \times n_q$ kernel matrix between query points.
        \item $K_{qs} = [k_{\phi}(x_{q,i}, x_{s,j})]_{i=1,j=1}^{n_q,n_s}$, the $n_q \times n_s$ kernel matrix between query and support points.
        \item $K_{sq} = K_{qs}^T$, the $n_s \times n_q$ kernel matrix between support and query points.
    \end{itemize}
    The posterior predictive covariance is then given by $\Sigma_{q|s} = K_{qq} - K_{qs}(K_{ss} + \sigma_n^2 I_{N_s})^{-1} K_{sq}$, where $\sigma_n^2$ is the observational noise variance and $I_{n_s}$ is the $n_s \times n_s$ identity matrix. The relevant term in our objective related to model complexity is $-\frac{1}{2}\log|\Sigma_{q|s}|$. By properties of determinants of positive definite block matrices (specifically, the joint prior covariance matrix over $f_s$ and $f_q$), it follows that $|\Sigma_{q|s}| \le |K_{qq}|$. Therefore, $\log |\Sigma_{q|s}| \le \log |K_{qq}|$, and minimizing $\log |\Sigma_{q|s}|$ as part of our objective is a different optimization target than directly minimizing $\log |K_{qq}|$. We hypothesize that this difference mitigates some of the detrimental effect of minimizing the log determinant of the \emph{prior} covariance (i.e., $\log|K_{qq}|$ or terms in $\log|K_{ss}+\sigma_n^2I_{n_s}|$), reducing our susceptibility to the failure mode studied by Ober et al.
    Notably, the outer meta-learning loop objective used by \citep{chen2023meta} is also based on the query set predictive posterior, suggesting they might also have benefited from this effect.

\textbf{Explicit Epistemic Uncertainty Optimization:} A more speculative reason relates to the nature of the objective. Feature collapse in deep kernel GPs affects the uncertainty of their latent function values. Therefore, optimizing for an objective that explicitly depends on the latent function uncertainty (i.e., epistemic uncertainty via $\Sigma_{q|s}$) may have encouraged the formation of features that avoid collapse. Specifically,  overconfident latent function predictions associated with feature collapse (i.e., near-singular $\Sigma_{q|s}$) would be heavily penalized by the data-fit component of the loss function (which involves $\Sigma_{q|s}^{-1}$), unless predictions were unrealistically perfect. This leads to a poor overall log probability. If we had trained on an objective incorporating observed (aleatoric) noise, a sufficiently large learned noise variance ($\sigma_{\text{aleatoric}}^2$) could stabilize the predictive covariance (i.e., in terms like $\Sigma_{q|s} + \sigma_{\text{aleatoric}}^2 I$), thereby masking the collapsed epistemic uncertainty $\Sigma_{q|s}$ and reducing the direct pressure on the feature extractor to maintain well-calibrated epistemic uncertainty.

\section{Implementation Details and Runtimes}
\label{app:implementation}
\subsection{Software and Hardware}
All models were implemented and experiments were conducted using Python \texttt{3.11.12}.
The primary deep learning library utilized was PyTorch \texttt{2.6.0+cu124} \citep{paszke2019pytorch}.
Gaussian Process functionalities were implemented using GPyTorch \texttt{1.14} \citep{gardner2018gpytorch}.
Other key libraries included NumPy \texttt{2.0.2}, SciPy \texttt{1.14.1}, and Scikit-learn \texttt{1.6.1}.
The experiments were predominantly run on Google Colaboratory (Colab).
The primary compute hardware utilized for experiments was a single \texttt{NVIDIA L4} GPU,equipped with approximately \texttt{23 GiB} of VRAM (23034 MiB). We acknowledge the use of Large Language Models (LLMs) to assist with generating boilerplate code, suggesting edits, and aiding debugging tasks during the development process- all strictly reviewed for correctness. 

\subsection{Baseline and Ablation Model Details}
\label{sec:baselinedetails}

\textbf{Convolutional Neural Network (CNN):}
    We implemented a CNN baseline by exactly replicating the ``Systems Identification'' branch architecture presented in \citep{qiu2023efficient}, selected because it represents a well-validated model for this specific dataset. The architecture consists of two layers: a convolutional layer employing $9 \times 9$ kernels, followed by a fully connected layer. Consistent with \citep{qiu2023efficient}, L2 regularization was applied to the convolutional filters to encourage smooth, center-surround-like receptive fields, while L1 regularization was applied to the weights of the fully connected layer to promote sparse feature usage for each output neuron. Hyperparameters were determined through grid search. The search space spanned: number of convolutional channels $\in \{8, 16, 24, 32\}$, L2 regularization coefficient $\in \{0, 0.1, 1.0, 10.0\}$, L1 regularization coefficient $\in \{0, 10^{-3}, 10^{-2}, 1/16, 0.1\}$, and learning rate $\in \{10^{-2}, 3 \times 10^{-3}, 10^{-3}, 3 \times 10^{-4}, 10^{-4}\}$. These ranges were based on the values used in the original work \citep{qiu2023efficient}. Each model variant was trained for $12$ epochs using the Adam optimizer ($\beta_1=0.99, \beta_2=0.999$) with a batch size of $32$. Training incorporated early stopping based on performance evaluated on a held-out validation/query dataset.
    
\textbf{Standard RBF Gaussian Process (GP):}
    The Radial Basis Function (RBF) GP baseline was implemented using the identical meta-adaptation script employed for our main model and ablation studies, omitting the feature extraction components. To ensure comparability, the same random seeds were utilized, acknowledging the sensitivity of non-parametric models to the specific conditioning data. GP hyperparameters (kernel lengthscale and variance) were optimized by maximizing the marginal likelihood. We initialized the lengthscale using the median heuristic and employed a broad prior variance ($100$) to avoid unduly constraining the lengthscale learning process. The model was trained for $300$ epochs using the Adam optimizer with a learning rate of $4 \times 10^{-3}$ and betas $(\beta_1=0.99, \beta_2=0.999)$. Early stopping was triggered based on the correlation coefficient measured on the held-out validation/query set.

\textbf{Linear-Nonlinear (LN) Model:}
    This baseline models neural responses $\mathbf{\hat{y}}$ based on flattened input images $\mathbf{x}$ (vectorized size $36 \times 32 = 1152$) according to the function:
    $$ \mathbf{\hat{y}} = g(\mathbf{W} \mathbf{x} + \mathbf{b}) $$
    In this formulation, $\mathbf{W}$ denotes the weight matrix mapping the input vector to the $86$ output neurons, $\mathbf{b}$ represents a bias vector, and $g$ is a pointwise nonlinearity. We employed the hyperbolic tangent ($tanh$) function for $f$. Model parameters ($\mathbf{W}$ and $\mathbf{b}$) were learned by minimizing the Mean Squared Error (MSE) loss between the predicted responses $\mathbf{\hat{y}}$ and the ground truth responses $\mathbf{y}$, utilizing the Adam optimizer. L2 regularization was applied to the weight matrix $\mathbf{W}$. Hyperparameters were selected via grid search, optimizing the learning rate (6 values logarithmically spaced within $[5 \times 10^{-4}, 10^{-1}]$) and the L2 regularization coefficient (14 values logarithmically spaced from $10^{-6}$ to $10^{1}$ and 0). The optimal hyperparameter combination was chosen and early stopping triggered based on the highest correlation coefficient achieved on a held-out validation set. The model was trained for $400$ epochs with a batch size of $20$, using Adam with betas $(\beta_1=0.99, \beta_2=0.999)$.

\textbf{Ablation 1: Identity Deep Kernel}
        This ablation serves to isolate the contribution of the meta-learned deep feature extractor. It removes the feature extractor entirely, directly applying the task-adaptive linear heads ($h_i$) to the raw input pixels $\mathbf{x}$. The output features from the linear head are then fed into the task-specific RBF GP layer. The task-adaptive parameters (linear head weights $W_i$ and GP hyperparameters $\theta_{gp,i}$) were optimized for each task using the exact same adaptation procedure, hyperparameters, seeds, and initialization strategies described in Section \ref{sec:taskadaptationprocedure}, ensuring consistency with the main 'Informed' model.

\textbf{Ablation 2: Random Deep Kernel}
        This ablation acts as a crucial control for the meta-learning process, testing the benefit of the theory-informed initialization against a random one, given the same fixed-extractor adaptation framework. It uses the same deep kernel architecture as the 'Informed' model (feature extractor $\phi$, task-adaptive linear heads $h_i$, RBF GP layer). However, the weights of the feature extractor $\phi$ were initialized randomly using the same scheme as the meta-learning initialization, but without loading the pre-trained weights. Crucially, mirroring the procedure for the 'Informed' model during task adaptation on the biological data, these randomly initialized feature extractor weights were kept frozen. Only the task-adaptive parameters (linear head weights $W_i$ and GP hyperparameters $\theta_{gp,i}$) were optimized using the exact same adaptation procedure, hyperparameters, seeds,  and initialization strategies described in Section \ref{sec:taskadaptationprocedure}. This ensures a direct comparison of the utility of the frozen meta-learned features versus frozen random features within our adaptation framework.

\subsection{Runtimes}
We did not maintain precise runtime logs during experimentation; the following estimates are based on file timestamps, run on a single NVIDIA L4 GPU.
The full meta-training process for our proposed model, including intermediate evaluations, required an estimated 3.5-4.5 hours. Generating the results presented for our model in Figure 3a (which involved evaluations across 10 training set sizes, for all 86 neurons) necessitated a total estimated runtime between 6 and 12 hours. The RBF GP baseline required total computational times comparable to our full informed model (~6-12 hours). While the RBF GP lacks neural network computations during adaptation, its runtime may be influenced by factors such as kernel matrix conditioning. The CNN baseline required approximately 5-10 minutes for final training, not counting the hyperparameter tuning time.
Runtimes for parametric and nonparametric models used here are not directly comparable. The CNN predicts responses for all 86 neurons in parallel, while the adaptation phase for the GP models treats each neuron as a separate task with a 1D output.

\section{Prototype Extraction and Interpretability Analysis}
\label{app:prototypes}
This section provides further details on the method used to extract and analyze prototype images, offering insight into the task-specific representations learned by the linear heads of our deep kernel Gaussian Process model. 

\subsection{Computation of Prototype Images}
\label{app:prototype_derivations}
\textit{For visual clarity, we opted to use the superscript $(i)$ here to denote task-indices, rather than the standard subscript $i$ throughout }

The "Prototype Image" interpretation method relies on analyzing how the task-specific linear heads $h^{(i)}$ (with parameters $W_i$), modify the geometry of the data representation learned by the shared feature extractor, $\phi$. We start by defining the pairwise Euclidean distances between input images $x_j$ and $x_k$ (vectorized) in the input space:
\[
D^{(x)}_{jk}\;=\;\bigl\lVert x_j - x_k \bigr\rVert_2 .
\]
Images are then passed through the shared feature extractor $\phi(\cdot)$:
\[
\phi_j \;=\;\phi\!\bigl(x_j\bigr).
\]
The pairwise distances are computed in this shared feature space:
\[
D^{(\phi)}_{jk}\;=\;\bigl\lVert \phi_j - \phi_k \bigr\rVert_2 .
\]
Subsequently, for each task $i$, the features are transformed by the task-specific linear head $h^{(i)}$:
\[
h^{(i)}_j \;=\;W_i \phi_j.
\]
The pairwise distances are computed again in this task-specific representation space:
\[
D^{(i)}_{jk}\;=\;\bigl\lVert h^{(i)}_j - h^{(i)}_k \bigr\rVert_2 .
\]
As a practical implementation detail, before computing the difference matrix $\Delta^{(i)}_{jk}$, the distance matrices $D^{(\phi)}_{jk}$ and $D^{(i)}_{jk}$ are normalized (by dividing by their respective maximum absolute values) to ensure they are on a comparable scale.

The core idea is to quantify how the task-specific head $h^{(i)}$ changes the relative distances between pairs of images compared to the shared feature space. This change is captured by the difference matrix $\Delta^{(i)}_{jk}$:
\[
\Delta^{(i)}_{jk}\;=\;D^{(\phi)}_{jk}\;-\;D^{(i)}_{jk}.
\]
This matrix indicates pairs of images whose relative distance decreased ($ \Delta^{(i}_{jk} > 0 $) or increased ($ \Delta^{(i)}_{jk} < 0 $) due to the task-specific transformation $h^{(i)}$.

To isolate the effect relative to the average change induced by the head for a given image $j$, we zero-center the difference matrix row-wise (i.e., for each fixed image $j$):
\[
\widetilde{\Delta}^{(i)}_{jk}\;=\;
\Delta^{(i)}_{jk}
-
\frac{1}{N}\sum_{k'=1}^{N}\Delta^{(i)}_{jk'},
\]
where $N$ is the total number of images considered in the analysis. For our result we use $N=750$.

Next, to relate these distance changes back to the input space, we compute a pixel-wise \emph{overlap function}, $O_{jk}(u)$, between each pair of images $x_j$ and $x_k$. This function measures the similarity of pixel intensities $a_j(u)$ and $a_k(u)$ at each pixel location $u$, using a Gaussian weighting:
\[
O_{jk}(u)\;=\;\exp\!\Bigl[-\tfrac{\bigl(a_j(u)-a_k(u)\bigr)^2}{2\sigma^2}\Bigr].
\]
Here, $\sigma$ is a bandwidth hyperparameter controlling the sensitivity to pixel intensity differences, and we use $\sigma=0.01$, but the method yielded similar results for other small values of $\sigma$ explored

For each image $j$ and each task $i$, we compute a \emph{per-image contribution} map, $C_j^{(i)}(u)$. This map represents, for image $j$, the average pixel-wise overlap with all other images $k$, weighted by the normalized change in pairwise distance $\widetilde{\Delta}^{(i)}_{jk}$ induced by the task-specific head $h^{(i)}$. The normalization ensures that the contribution is scaled relative to the total magnitude of distance changes involving image $j$:
\[
C_j^{(i)}(u)\;=\;
\frac{\displaystyle\sum_{j\neq k}\,
\widetilde{\Delta}^{(i)}_{jk}\,
O_{jk}(u)}
{\displaystyle\sum_{j\neq k}\bigl|\widetilde{\Delta}^{(i)}_{jk}\bigr| + \epsilon},
\]
where $\epsilon$ is a small constant (e.g., $10^{-8}$) added for numerical stability.

Finally, the overall \emph{prototype image} $P^{(i)}(u)$ for task $i$ is obtained by averaging these per-image contributions across all $N$ images:
\[
P^{(i)}(u)\;=\;\frac{1}{N}\sum_{j=1}^{N}C_j^{(i)}(u).
\]
This resulting image $P^{(i)}(u)$ can be interpreted as visualizing the average input pattern that leads to the largest relative changes in representation distances specific to task $i$.

\subsection{Additional Visualizations}
\label{app:prototype_visuals}

Here we extend the visualizations presented in Figures 4a and 4c.

\begin{figure}[h!]
    \centering
        \includegraphics[width=\textwidth]{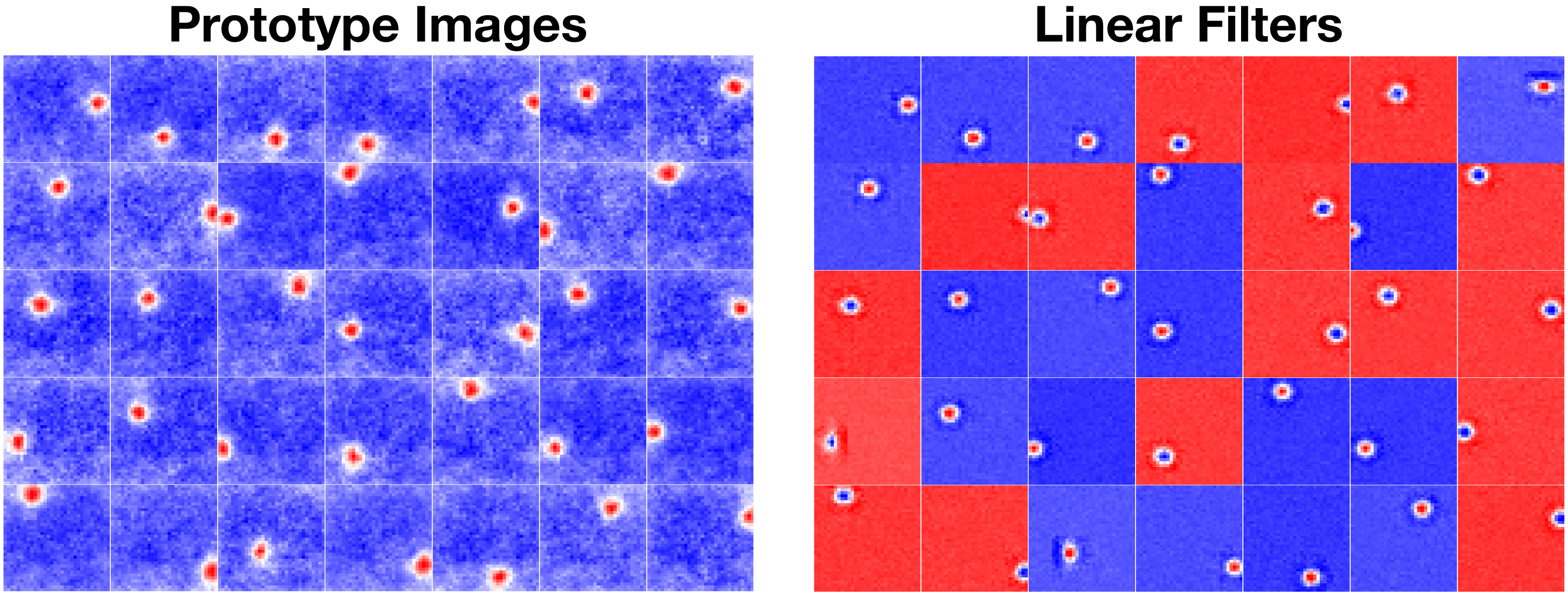} 

    \caption{Additional examples comparing extracted prototypes $P^{(i)}(u)$ to ground truth linear filters for synthetic data.}
    \label{fig:app_synth_prototypes}
\end{figure}

\textbf{Extended 4a Synthetic Data Prototypes:} Figure~\ref{fig:app_synth_prototypes} presents additional examples comparing the extracted prototype images $P^{(i)}(u)$ with the ground truth linear filters for the synthetic dataset experiments. 

\begin{figure}[H]
    \centering
    \includegraphics[width=0.6\textwidth]{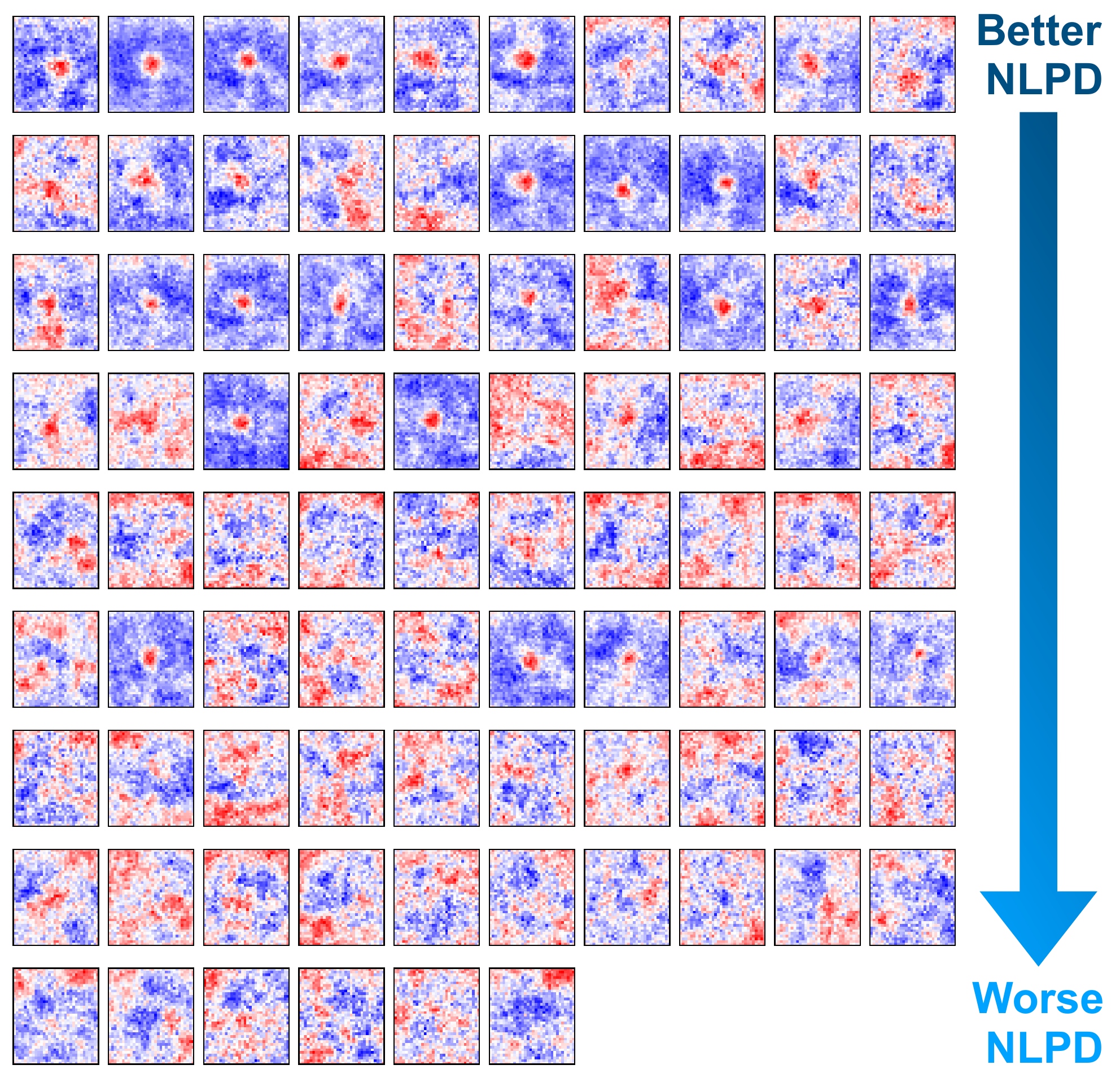} 
    \caption{Prototype images $P^{(i)}(u)$ for a broader selection of biological neurons, ordered by performance (NLPD) showing trends consistent with those of figure 4c}
    \label{fig:app_bio_prototypes_wide}
\end{figure}

\textbf{Extended 4c Biological Neuron Prototypes:} Figure~\ref{fig:app_bio_prototypes_wide} displays prototype images $P^{(i)}(u)$ for a more extensive set of biological neurons beyond the best and worst 25 examples shown in Figure 4c.

\section{Bayesian Model Comparison and Optimality Validation}
\label{sec:bmc_ablations}

\subsection{Ablation Study on Bayesian Model Comparison}
To  confirm that our method is specifically sensitive to the theory's structure, we applied our ablation conditions on our Theory Informed model for the Bayesian Model comparison analysis. Here, we report the Pearson correlation coefficients obtained by implementing the exact same "identification of optimality level" protocol using the "Random", "Identity", and "Task Ablation" conditions. 

\begin{figure}[H]
    \centering
    \includegraphics[width=\textwidth]{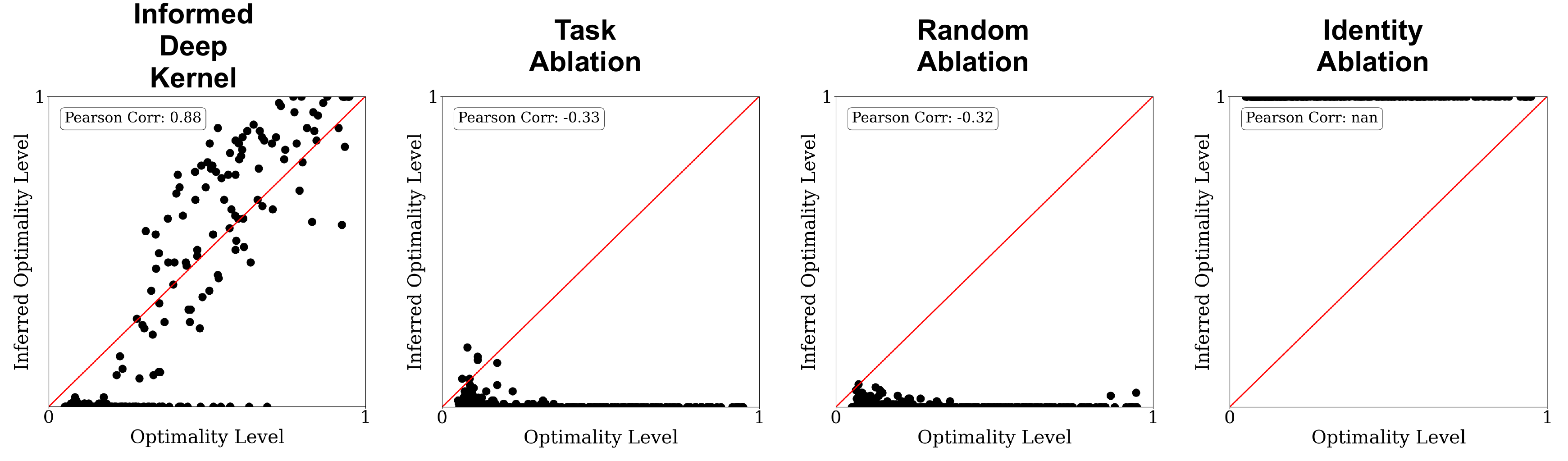} 
    \caption{Ablation Study on Bayesian Model Comparison shows Theory Informed Features indispensable}
    \label{fig:5a Ablation}
\end{figure}

When using the unstructured prior, the inferred optimality level is either anti-correlated (Pearson corr. = -0.32, Pearson corr. =-0.33 ) with ground truth, or simply not sensitive at all (Figure \ref{fig:5a Ablation}). This confirms that the model comparison capability is driven specifically by the theory/data match.

\subsection{Mixture Kernel Formulation}
\label{app:mixture_kernel}

To smoothly interpolate between a full explanation by the normative theory and a null hypothesis, we define a mixture kernel $K_{\beta}$ as a convex combination of the theory-informed kernel $K_{\text{TIK}}$ and a standard Radial Basis Function (RBF) kernel $K_{\text{RBF}}$:
\[
K_\beta(x,x') = \beta K(x',x)_{\text{TIK}} + (1-\beta)K(x',x)_{\text{RBF}}
\]
Here, $\beta \in [0, 1]$ is a mixing coefficient that represents the degree of belief allocated to the theory-informed model. $K_{\text{TIK}}$ encodes the prior assumptions derived from the specific normative theory being tested (e.g., efficient coding leading to center-surround receptive fields). The RBF kernel, defined as $K_{\text{RBF}}(x, x') = \theta_0^2 \exp\left(-\frac{\|x - x'\|^2}{2\ell^2}\right)$ with outputscale $\theta_0$ and lengthscale $\ell$, serves as the null model.

Our rationale for using the RBF kernel as the null model is that it represents a flexible, general-purpose prior over smooth functions \citep{williams2006gaussian}, and that the Deep Kernel model itself is an RBF GP operating on top of a learned feature space. The RBF makes minimal structural assumptions about the data-generating process beyond smoothness, controlled by its hyperparameters ($\theta_0, \ell$). Therefore, it provides a reasonable baseline against which the specific structural assumptions encoded in $K_{\text{TIK}}$ can be compared. A higher inferred $\beta^*$ indicates that the specific structure imposed by the normative theory provides a better explanation for the data than the generic smoothness assumption of the RBF kernel.

\subsection{Inference of Optimality Level (\texorpdfstring{$\beta^*$}{beta*})}

\label{app:beta_inference}

We infer the optimal mixing coefficient $\beta^*$ by maximizing the marginal likelihood of the Gaussian Process model using the mixture kernel $K_\beta$. The marginal likelihood $P(Y|X,\beta)$ represents the probability of observing the neural responses $Y$ given the stimuli $X$ and the specific mixed prior defined by $\beta$. The optimal $\beta^*$ is found by:
\[
\beta^{*} = \arg \max_{\beta \in [0, 1]} P(Y|X,\beta) \quad 
\]
This maximization is performed by evaluating the marginal likelihood across a discretized grid of $\beta$ values. Specifically, we evaluated $P(Y|X,\beta)$ for 100 evenly spaced values of $\beta$ ranging from 0 to 1.

Crucially, before evaluating the marginal likelihood across the $\beta$ grid, other hyperparameters of the model are optimized and then frozen. This includes:
\begin{enumerate}
    \item Task Adaptive parameters of the informed kernel $K_{\text{TIK}}$, (i.e. linear head weights and GP layer hyperparameters).
    \item Hyperparameters of the RBF kernel $K_{\text{RBF}}$ (i.e. outputscale $\theta_0$, lengthscale $\ell$, and noise).
\end{enumerate}
These parameters are learned by maximizing the marginal likelihood of their respective base models  Once optimized, these parameters are held fixed during the grid search for $\beta^*$.  This ensures that the comparison focuses specifically on the contribution of the informed structure versus the null structure, as interpolated by $\beta$ \citep{mlynarski2021statistical}.

\subsection{Synthetic Validation Details}
\label{app:synthetic_validation}

To validate our Bayesian model comparison approach for inferring optimality, we first applied it to synthetic data where the "ground truth" level of optimality could be controlled and measured independently.

\paragraph{Generating Controlled Optimality Receptive Fields (RFs)}
\label{app:generating_suboptimal}

We heuristically generated synthetic receptive fields (RFs) with varying degrees of optimality using a controlled noise-injection process. The procedure involved the following steps:

\begin{enumerate}
    \item \textbf{Initial Optimal RFs:} We started with a set of 20 archetypal synthetic "optimal" RFs.

    \item \textbf{Generating "Anti-Optimal" Noise:} To perturb these RFs away from optimality in a structured manner, we generated a basis for noise that is explicitly orthogonal to any possible center-surround receptive fields like ours. This was achieved by:
    \begin{itemize}
        \item Taking a larger reference set of 4000 theoretically "optimal" center-surround RFs (distinct from the initial 20 RFs and any meta-training data). Let this set be $J_{RF} \in \mathbb{R}^{4000 \times H \times W}$. We also considered the transposed versions $J_{RF}^T$.
        \item Flattening these RFs into vectors $J_{\text{flat}} \in \mathbb{R}^{8000 \times (H \cdot W)}$.
        \item Performing Singular Value Decomposition (SVD) on $J_{\text{flat}}$: $J_{\text{flat}} = U S V^T$.
        \item Determining the effective rank $r$ of the optimal RF subspace by thresholding the singular values $S$. We used a threshold ratio of $0.1$ relative to the maximum singular value.
        \item Identifying the subspace orthogonal to the dominant $r$ principal components (columns of $V$). The projection matrix onto the optimal subspace is $P = V_r V_r^T$, where $V_r$ contains the first $r$ columns of $V$. The projection onto the orthogonal subspace is $P_{\perp} = I - P$.
        \item Generating noise images by projecting natural images ($N_{\text{img}}$) onto this orthogonal subspace: $\text{Noise}_{\text{img}} = P_{\perp} N_{\text{img}}$. These images represent patterns or structures that are explicitly \emph{anti} optimal. Figure \ref{fig:noise} shows examples of such generated noise components.
    \end{itemize}

    \begin{figure}[h!]
        \centering
        \includegraphics[width=0.8\textwidth]{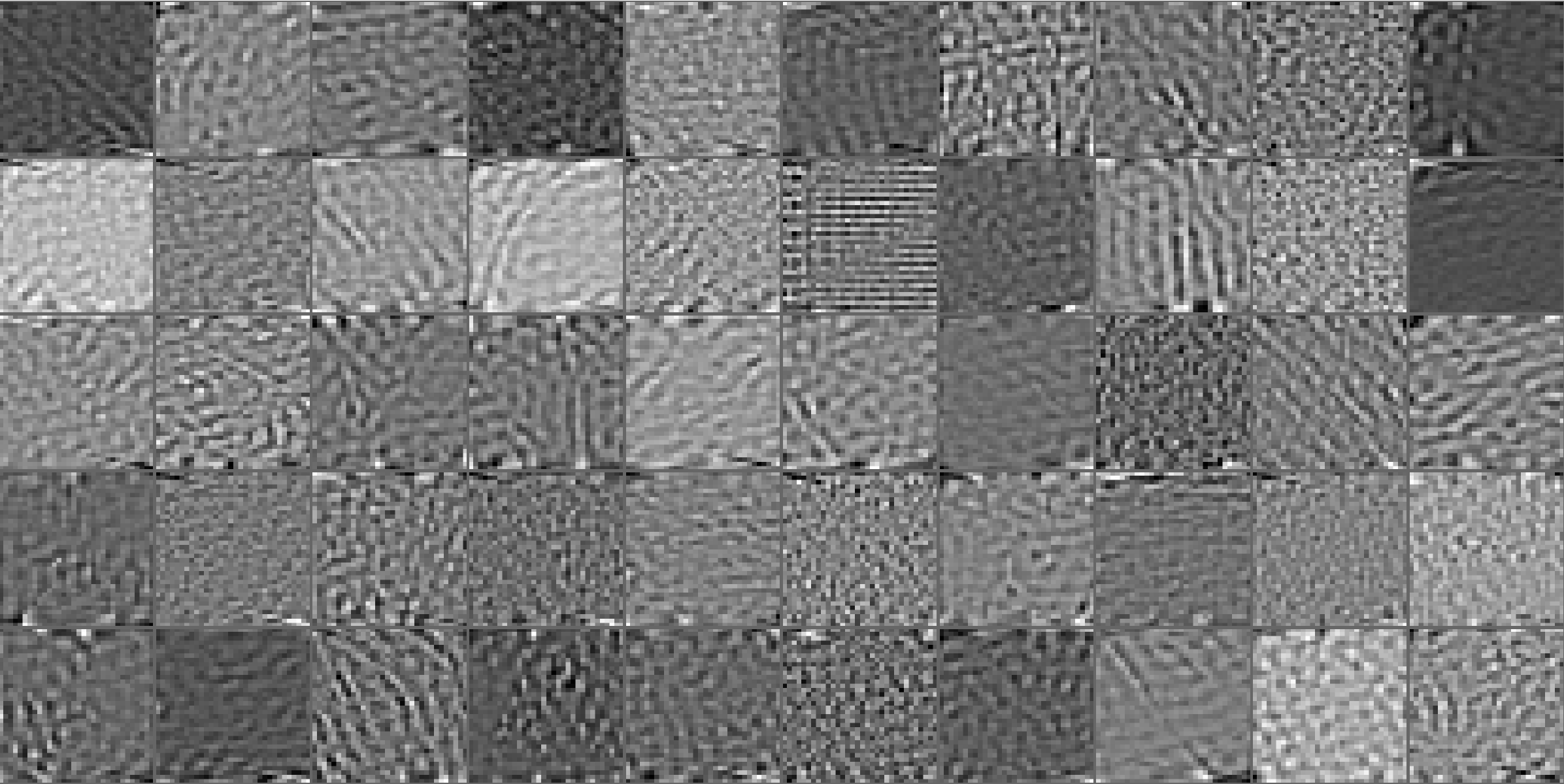} 
        \caption{Example "Noise" images used for generating suboptimal RFs. These images were generated by finding the component of natural images orthogonal to the subspace spanned by a large set of optimal center-surround receptive fields.}
        \label{fig:noise}
    \end{figure}

    \item \textbf{Perturbation Process:} We simulated a random walk for each initial optimal RF over $T$ timesteps (Here, T=600). At each timestep $t$:
    \begin{itemize}
        \item We sample a random "anti-optimal" noise image.
        \item This noise image was scaled by a random Gaussian scalar $z \sim \mathcal{N}(0,\text{scale})$, where scale = 0.01 for our purposes.
        \item The scaled noise was added to the RF from timestep $t-1$: $RF_t = RF_{t-1} + z \cdot \text{Noise}_{\text{img}}$.
        \item The resulting $RF_t$ was normalized by subtracting its mean and dividing by its L2 norm: $RF_t = (RF_t - \overline{RF}_t) / \|RF_t\|_2$. This process gradually pushes the RF away from its initial optimal structure .
    \end{itemize}
    
    \item \textbf{Subsampling Trajectories:} The full perturbation process generates long trajectories (T=600 steps) for each initial RF. To create a dataset with distinct, relatively stable levels of suboptimality, we subsampled these trajectories. For each RF's trajectory of ground truth $R^2$ values (Sec. \ref{app:dog_fitting}), we selected a smaller number of points (20). The selection aimed to capture points along a smooth, approximately linear decay in $R^2$ by minimizing the residual from a global linear fit within predefined index windows. This ensures the final synthetic dataset contains RFs spanning a range of controlled optimality levels, while avoiding pathological points, for instance when the DoG model fails to fit and produces a nonsense $R^2$ value.
\end{enumerate}

\paragraph{Ground Truth Optimality Quantification: $R^2$ of DoG}
\label{app:dog_fitting}

To obtain an independent, "ground truth" measure of optimality for the generated synthetic RFs, we used the standard approach of fitting a Difference of Gaussians (DoG) model to each RF and calculating the coefficient of determination ($R^2$). This is motivated by the theoretical understanding that optimal RFs under certain efficient coding assumptions often resemble DoG functions \citep{atick1992does}. 

\begin{enumerate}
    \item \textbf{DoG Model:} The DoG model is defined as the difference of two concentric Gaussian functions with different amplitudes and standard deviations:
    \[
    \text{DoG}(x, y | \theta) = A_c \exp\left(-\frac{(x-x_0)^2 + (y-y_0)^2}{2\sigma_c^2}\right) - A_s \exp\left(-\frac{(x-x_0)^2 + (y-y_0)^2}{2\sigma_s^2}\right)
    \]
    The parameter vector is $\theta = \{A_c, A_s, x_0, y_0, \sigma_c, \sigma_s\}$, representing the amplitudes of the center and surround Gaussians, their common center $(x_0, y_0)$, and their standard deviations ($\sigma_c, \sigma_s$). 
    
    \item \textbf{Fitting Procedure:} For each synthetic RF (dimensions $D1 \times D2$), we fitted the DoG model by minimizing the mean squared error between the flattened RF pixel values ($RF_{flat}$) and the DoG model output evaluated on a corresponding grid ($DoG_{flat}(\theta)$):
    \[
    \hat{\theta} = \arg \min_{\theta} \frac{1}{D1 \cdot D2} \| \text{RF}_{\text{flat}} - \text{DoG}_{\text{flat}}(\theta) \|^2
    \]

    \item \textbf{Initialization:} Parameter initialization significantly impacts DoG fitting. We used the following strategy:
    \begin{itemize}
        \item The scale of the center, $A_c$ was initialized to the maximum value of the RF.
        \item The scale of the Surround, $A_s$ was initialized to $0.5 \times A_c$.
        \item $\log \sigma_c$ and $\log \sigma_s$ were initialized to fixed values corresponding to $\sigma_c \approx 3.0$ and $\sigma_s \approx 6.0$.
        \item The center location $(x_0, y_0)$ was initialized using a localized center-of-mass heuristic. This involved calculating 1D summed projections of the RF along rows and columns. A 1D convolution with a sliding window was applied to these projections to find the region of maximal energy. The center of mass was then computed within this window. To improve robustness, this process was repeated for multiple window sizes . The DoG fit was performed for each resulting $(x_0, y_0)$ initialization, and the fit yielding the highest final $R^2$ was selected.
    \end{itemize}

    \item \textbf{$R^2$ Calculation:} After fitting, the goodness-of-fit was quantified using the coefficient of determination ($R^2$):
    \[
    R^2 = 1 - \frac{\sum_{i=1}^{D1 \cdot D2} (\text{RF}_{\text{flat}, i} - \text{DoG}_{\text{flat}, i}(\hat{\theta}))^2}{\sum_{i=1}^{D1 \cdot D2} (\text{RF}_{\text{flat}, i} - \overline{\text{RF}}_{\text{flat}})^2}
    \]
    where $\overline{\text{RF}}_{\text{flat}}$ is the mean value of the RF pixels. This $R^2$ value serves as the "ground truth" optimality measure for each synthetic RF, against which our inferred $\beta^*$ values were compared (in Figure 5a of the main text).
\end{enumerate}

\newpage
\section{Global Hyperparameter Table}

\begin{table}[h!]
\centering
\caption{Summary of Key Hyperparameters}
\label{tab:full_hyperparameters}
\small 
\begin{tabular}{lll}
\toprule
\textbf{Parameter Category} & \textbf{Hyperparameter} & \textbf{Value / Setting} \\
\midrule

\multirow{9}{*}{\textbf{Meta-Training}}
& Optimizer & Adam \\
& Total Epochs ($E$) & 7 \\
& Meta Batch Size ($B$, num tasks) & 10 \\
& Total Meta-Train Tasks & 490 (49 batches $\times$ 10 tasks) \\
& Task Data Points (Support + Query) & 1452 \\
& Support/Query Split & 5\% Support / 95\% Query (resampled each epoch) \\
& Inner Adaptation Steps ($n_{\text{inner steps}}$) & 17 \\
& Inner LR - Linear Head ($\alpha_{\text{linear}}$) & $4 \times 10^{-4}$ (0.0004) \\
& Inner LR - GP Params ($\alpha_{\text{GP}}$) & $2 \times 10^{-3}$ (0.002) \\
& Outer Meta-Update Steps ($n_{\text{outer steps}}$) & 3 \\
& Outer LR ($\alpha_{\text{outer}}$) & $4 \times 10^{-4}$ (0.0004) \\
& Adam Betas ($\beta_1, \beta_2$) & (0.5,0.5) \\
& Gradient Clipping (Feature Extractor Norm) & 1.0 \\
& Initial LR Scaling (First Epoch) & 1/10th \\
& GP Noise $\sigma_{\eta}$ (Synthetic Tasks) &  $10^{-4}$ \\
& Data Type & torch.double \\
\midrule

\multirow{7}{*}{\textbf{Task Adaptation}}
& Adapted Components & Linear Heads ($h_i$), GP Hyperparameters ($\theta_{gp,i}$) \\
& Frozen Components & Feature Extractor ($\phi$, both Informed and Random) \\
& Optimizer & Adam \\
& Epochs & 300 \\
& Base Learning Rate (for GP Params) & $4 \times 10^{-3}$ \\
& Linear Head LR Scale & $10^{-2}$ (Effective LR: $4 \times 10^{-5}$) \\
& Adam Betas ($\beta_1, \beta_2$) & (0.99, 0.999) \\
& L1 Sparsity (Linear Heads) & $10^{-2}$ \\
& GP Lengthscale Prior & Gaussian(mean=Median Init, var=0.01) \\
& GP Lengthscale Init & Median Heuristic  \\
& GP Noise $\sigma_{\eta}$ Init (Biological) & Standard GPyTorch Initialization \\
& GP Noise $\sigma_{\eta}$ Init (Synthetic Test) & $10^{-4}$ \\
\midrule

\multirow{7}{*}{\textbf{CNN Baseline}}
& Architecture & Qiu et al. \citep{qiu2023efficient} "SI Branch" \\
& Optimizer & Adam \\
& Adam Betas ($\beta_1, \beta_2$) & (0.99, 0.999) \\
& Epochs & 12  \\
& Batch Size & 32 \\
& Hyperparameter Selection & Grid Search (Val. Correlation) \\
& Grid: Conv Channels & \{8, 16, 24, 32\} \\
& Grid: Conv L2 Reg & \{0, 0.1, 1.0, 10.0\} \\
& Grid: FC L1 Reg & \{0, $10^{-3}$, $10^{-2}$, 1/16, 0.1\} \\
& Grid: Learning Rate & \{$10^{-2}$, $3 \times 10^{-3}$, $10^{-3}$, $3 \times 10^{-4}$, $10^{-4}$\} \\
\midrule

\multirow{6}{*}{\textbf{RBF GP Baseline}}
& Adapted Components & GP Hyperparameters ($\theta_{gp,i}$) \\
& Optimizer & Adam \\
& Epochs & 300 (with Early Stopping on Val. Correlation) \\
& Learning Rate & $4 \times 10^{-3}$ \\
& Adam Betas ($\beta_1, \beta_2$) & (0.99, 0.999) \\
& Lengthscale Init & Median Heuristic \\
& Lengthscale Prior Variance & 100 (Broad) \\
\midrule

\multirow{8}{*}{\textbf{LN Baseline}}
& Non-linearity ($g$) & tanh \\
& Optimizer & Adam \\
& Adam Betas ($\beta_1, \beta_2$) & (0.99, 0.999) \\
& Epochs & 400 \\
& Batch Size & 20 \\
& Hyperparameter Selection & Grid Search (Val. Correlation) \\
& Grid: Learning Rate & 6 values log-spaced [$5 \times 10^{-4}$, $10^{-1}$] \\
& Grid: L2 Reg (Weights $\mathbf{W}$) & 14 values log-spaced [$10^{-6}$, $10^{1}$] + [$0$] \\

\bottomrule
\end{tabular}
\end{table}

%% file: neuraltaskkernels.bbl
\begin{thebibliography}{75}
\providecommand{\natexlab}[1]{#1}
\providecommand{\url}[1]{\texttt{#1}}
\expandafter\ifx\csname urlstyle\endcsname\relax
  \providecommand{\doi}[1]{doi: #1}\else
  \providecommand{\doi}{doi: \begingroup \urlstyle{rm}\Url}\fi

\bibitem[Atick(1992)]{atick1992could}
Joseph~J Atick.
\newblock Could information theory provide an ecological theory of sensory
  processing?
\newblock \emph{Network: Computation in neural systems}, 3\penalty0
  (2):\penalty0 213--251, 1992.

\bibitem[Atick \& Redlich(1992)Atick and Redlich]{atick1992does}
Joseph~J Atick and A~Norman Redlich.
\newblock What does the retina know about natural scenes?
\newblock \emph{Neural computation}, 4\penalty0 (2):\penalty0 196--210, 1992.

\bibitem[Attneave(1954)]{attneave1954some}
Fred Attneave.
\newblock Some informational aspects of visual perception.
\newblock \emph{Psychological review}, 61\penalty0 (3):\penalty0 183, 1954.

\bibitem[Barlow et~al.(1961)]{barlow1961possible}
Horace~B Barlow et~al.
\newblock Possible principles underlying the transformation of sensory
  messages.
\newblock \emph{Sensory communication}, 1\penalty0 (01):\penalty0 217--233,
  1961.

\bibitem[Bengio et~al.(2013)Bengio, Courville, and
  Vincent]{bengio2013representation}
Yoshua Bengio, Aaron Courville, and Pascal Vincent.
\newblock Representation learning: A review and new perspectives.
\newblock \emph{IEEE transactions on pattern analysis and machine
  intelligence}, 35\penalty0 (8):\penalty0 1798--1828, 2013.

\bibitem[Bernardo \& Smith(2009)Bernardo and Smith]{bernardo2009bayesian}
Jos{\'e}~M Bernardo and Adrian~FM Smith.
\newblock \emph{Bayesian theory}, volume 405.
\newblock John Wiley \& Sons, 2009.

\bibitem[Bialek(2012)]{bialek2012biophysics}
William Bialek.
\newblock \emph{Biophysics: searching for principles}.
\newblock Princeton University Press, 2012.

\bibitem[Bittner et~al.(2021)Bittner, Palmigiano, Piet, Duan, Brody, Miller,
  and Cunningham]{bittner2021interrogating}
Sean~R Bittner, Agostina Palmigiano, Alex~T Piet, Chunyu~A Duan, Carlos~D
  Brody, Kenneth~D Miller, and John Cunningham.
\newblock Interrogating theoretical models of neural computation with emergent
  property inference.
\newblock \emph{Elife}, 10:\penalty0 e56265, 2021.

\bibitem[Breiman(2001)]{breiman2001statistical}
Leo Breiman.
\newblock Statistical modeling: The two cultures (with comments and a rejoinder
  by the author).
\newblock \emph{Statistical science}, 16\penalty0 (3):\penalty0 199--231, 2001.

\bibitem[Cadena et~al.(2019)Cadena, Denfield, Walker, Gatys, Tolias, Bethge,
  and Ecker]{cadena2019deep}
Santiago~A Cadena, George~H Denfield, Edgar~Y Walker, Leon~A Gatys, Andreas~S
  Tolias, Matthias Bethge, and Alexander~S Ecker.
\newblock Deep convolutional models improve predictions of macaque v1 responses
  to natural images.
\newblock \emph{PLoS computational biology}, 15\penalty0 (4):\penalty0
  e1006897, 2019.

\bibitem[Calandra et~al.(2016)Calandra, Peters, Rasmussen, and
  Deisenroth]{calandra2016manifold}
Roberto Calandra, Jan Peters, Carl~Edward Rasmussen, and Marc~Peter Deisenroth.
\newblock Manifold gaussian processes for regression.
\newblock In \emph{2016 International joint conference on neural networks
  (IJCNN)}, pp.\  3338--3345. IEEE, 2016.

\bibitem[Caruana(1997)]{caruana1997multitask}
Rich Caruana.
\newblock Multitask learning.
\newblock \emph{Machine learning}, 28:\penalty0 41--75, 1997.

\bibitem[Chen et~al.(2023)Chen, Tripp, and Hern{\'a}ndez-Lobato]{chen2023meta}
Wenlin Chen, Austin Tripp, and Jos{\'e}~Miguel Hern{\'a}ndez-Lobato.
\newblock Meta-learning adaptive deep kernel gaussian processes for molecular
  property prediction.
\newblock In \emph{The Eleventh International Conference on Learning
  Representations}, 2023.

\bibitem[Cowley et~al.(2017)Cowley, Williamson, Clemens, Smith, and
  Yu]{cowley2017adaptive}
Benjamin Cowley, Ryan Williamson, Katerina Clemens, Matthew Smith, and Byron~M
  Yu.
\newblock Adaptive stimulus selection for optimizing neural population
  responses.
\newblock \emph{Advances in neural information processing systems}, 30, 2017.

\bibitem[El-Gaby et~al.(2024)El-Gaby, Harris, Whittington, Dorrell, Bhomick,
  Walton, Akam, and Behrens]{el2024cellular}
Mohamady El-Gaby, Adam~Loyd Harris, James~CR Whittington, William Dorrell, Arya
  Bhomick, Mark~E Walton, Thomas Akam, and Timothy~EJ Behrens.
\newblock A cellular basis for mapping behavioural structure.
\newblock \emph{Nature}, 636\penalty0 (8043):\penalty0 671--680, 2024.

\bibitem[Frazier(2018)]{frazier2018tutorial}
Peter~I Frazier.
\newblock A tutorial on bayesian optimization.
\newblock \emph{arXiv preprint arXiv:1807.02811}, 2018.

\bibitem[Fumero et~al.(2023)Fumero, Wenzel, Zancato, Achille, Rodol{\`a},
  Soatto, Sch{\"o}lkopf, and Locatello]{fumero2023leveraging}
Marco Fumero, Florian Wenzel, Luca Zancato, Alessandro Achille, Emanuele
  Rodol{\`a}, Stefano Soatto, Bernhard Sch{\"o}lkopf, and Francesco Locatello.
\newblock Leveraging sparse and shared feature activations for disentangled
  representation learning.
\newblock \emph{Advances in Neural Information Processing Systems},
  36:\penalty0 27682--27698, 2023.

\bibitem[Gardner et~al.(2018)Gardner, Pleiss, Weinberger, Bindel, and
  Wilson]{gardner2018gpytorch}
Jacob Gardner, Geoff Pleiss, Kilian~Q Weinberger, David Bindel, and Andrew~G
  Wilson.
\newblock Gpytorch: Blackbox matrix-matrix gaussian process inference with gpu
  acceleration.
\newblock \emph{Advances in neural information processing systems}, 31, 2018.

\bibitem[Gelman et~al.(1995)Gelman, Carlin, Stern, and
  Rubin]{gelman1995bayesian}
Andrew Gelman, John~B Carlin, Hal~S Stern, and Donald~B Rubin.
\newblock \emph{Bayesian data analysis}.
\newblock Chapman and Hall/CRC, 1995.

\bibitem[Geng et~al.(2021)Geng, Zhang, Bai, Wang, and Lin]{geng2021training}
Zhengyang Geng, Xin-Yu Zhang, Shaojie Bai, Yisen Wang, and Zhouchen Lin.
\newblock On training implicit models.
\newblock \emph{Advances in Neural Information Processing Systems},
  34:\penalty0 24247--24260, 2021.

\bibitem[Goldin et~al.(2023)Goldin, Virgili, and Chalk]{goldin2023scalable}
Mat{\'\i}as~A Goldin, Samuele Virgili, and Matthew Chalk.
\newblock Scalable gaussian process inference of neural responses to natural
  images.
\newblock \emph{Proceedings of the National Academy of Sciences}, 120\penalty0
  (34):\penalty0 e2301150120, 2023.

\bibitem[H{\"u}llermeier \& Waegeman(2021)H{\"u}llermeier and
  Waegeman]{hullermeier2021aleatoric}
Eyke H{\"u}llermeier and Willem Waegeman.
\newblock Aleatoric and epistemic uncertainty in machine learning: An
  introduction to concepts and methods.
\newblock \emph{Machine learning}, 110\penalty0 (3):\penalty0 457--506, 2021.

\bibitem[Hyv{\"a}rinen et~al.(2009)Hyv{\"a}rinen, Hurri, and
  Hoyer]{hyvarinen2009natural}
Aapo Hyv{\"a}rinen, Jarmo Hurri, and Patrick~O Hoyer.
\newblock \emph{Natural image statistics: A probabilistic approach to early
  computational vision.}, volume~39.
\newblock Springer Science \& Business Media, 2009.

\bibitem[Jaynes(2003)]{jaynes2003probability}
Edwin~T Jaynes.
\newblock \emph{Probability theory: The logic of science}.
\newblock Cambridge university press, 2003.

\bibitem[Karklin \& Lewicki(2009)Karklin and Lewicki]{karklin2009emergence}
Yan Karklin and Michael~S Lewicki.
\newblock Emergence of complex cell properties by learning to generalize in
  natural scenes.
\newblock \emph{Nature}, 457\penalty0 (7225):\penalty0 83--86, 2009.

\bibitem[Karklin \& Simoncelli(2011)Karklin and
  Simoncelli]{karklin2011efficient}
Yan Karklin and Eero Simoncelli.
\newblock Efficient coding of natural images with a population of noisy
  linear-nonlinear neurons.
\newblock \emph{Advances in neural information processing systems}, 24, 2011.

\bibitem[Karniadakis et~al.(2021)Karniadakis, Kevrekidis, Lu, Perdikaris, Wang,
  and Yang]{karniadakis2021physics}
George~Em Karniadakis, Ioannis~G Kevrekidis, Lu~Lu, Paris Perdikaris, Sifan
  Wang, and Liu Yang.
\newblock Physics-informed machine learning.
\newblock \emph{Nature Reviews Physics}, 3\penalty0 (6):\penalty0 422--440,
  2021.

\bibitem[Kass \& Raftery(1995)Kass and Raftery]{kass1995bayes}
Robert~E Kass and Adrian~E Raftery.
\newblock Bayes factors.
\newblock \emph{Journal of the american statistical association}, 90\penalty0
  (430):\penalty0 773--795, 1995.

\bibitem[Kell et~al.(2018)Kell, Yamins, Shook, Norman-Haignere, and
  McDermott]{kell2018task}
Alexander~JE Kell, Daniel~LK Yamins, Erica~N Shook, Sam~V Norman-Haignere, and
  Josh~H McDermott.
\newblock A task-optimized neural network replicates human auditory behavior,
  predicts brain responses, and reveals a cortical processing hierarchy.
\newblock \emph{Neuron}, 98\penalty0 (3):\penalty0 630--644, 2018.

\bibitem[Kingma \& Ba(2014)Kingma and Ba]{kingma2014adam}
Diederik~P Kingma and Jimmy Ba.
\newblock Adam: A method for stochastic optimization.
\newblock \emph{arXiv preprint arXiv:1412.6980}, 2014.

\bibitem[Kobalczyk \& van~der Schaar(2025)Kobalczyk and van~der
  Schaar]{kobalczyk2025towards}
Kasia Kobalczyk and Mihaela van~der Schaar.
\newblock Towards automated knowledge integration from human-interpretable
  representations.
\newblock In \emph{The Thirteenth International Conference on Learning
  Representations}, 2025.

\bibitem[K{\"o}rding \& Wolpert(2006)K{\"o}rding and
  Wolpert]{kording2006bayesian}
Konrad~P K{\"o}rding and Daniel~M Wolpert.
\newblock Bayesian decision theory in sensorimotor control.
\newblock \emph{Trends in cognitive sciences}, 10\penalty0 (7):\penalty0
  319--326, 2006.

\bibitem[Lake et~al.(2017)Lake, Ullman, Tenenbaum, and
  Gershman]{lake2017building}
Brenden~M Lake, Tomer~D Ullman, Joshua~B Tenenbaum, and Samuel~J Gershman.
\newblock Building machines that learn and think like people.
\newblock \emph{Behavioral and brain sciences}, 40:\penalty0 e253, 2017.

\bibitem[Lewicki(2002)]{lewicki2002efficient}
Michael~S Lewicki.
\newblock Efficient coding of natural sounds.
\newblock \emph{Nature neuroscience}, 5\penalty0 (4):\penalty0 356--363, 2002.

\bibitem[Linderman \& Gershman(2017)Linderman and Gershman]{linderman2017using}
Scott~W Linderman and Samuel~J Gershman.
\newblock Using computational theory to constrain statistical models of neural
  data.
\newblock \emph{Current opinion in neurobiology}, 46:\penalty0 14--24, 2017.

\bibitem[Liu et~al.(2020)Liu, Lin, Padhy, Tran, Bedrax~Weiss, and
  Lakshminarayanan]{liu2020simple}
Jeremiah Liu, Zi~Lin, Shreyas Padhy, Dustin Tran, Tania Bedrax~Weiss, and
  Balaji Lakshminarayanan.
\newblock Simple and principled uncertainty estimation with deterministic deep
  learning via distance awareness.
\newblock \emph{Advances in neural information processing systems},
  33:\penalty0 7498--7512, 2020.

\bibitem[Lotfi et~al.(2022)Lotfi, Izmailov, Benton, Goldblum, and
  Wilson]{lotfi2022bayesian}
Sanae Lotfi, Pavel Izmailov, Gregory Benton, Micah Goldblum, and Andrew~Gordon
  Wilson.
\newblock Bayesian model selection, the marginal likelihood, and
  generalization.
\newblock In \emph{International Conference on Machine Learning}, pp.\
  14223--14247. PMLR, 2022.

\bibitem[MacKay(1992)]{mackay1992bayesian}
David~JC MacKay.
\newblock Bayesian interpolation.
\newblock \emph{Neural computation}, 4\penalty0 (3):\penalty0 415--447, 1992.

\bibitem[MacKay(2003)]{mackay2003information}
David~JC MacKay.
\newblock \emph{Information theory, inference and learning algorithms}.
\newblock Cambridge university press, 2003.

\bibitem[McCoy et~al.(2020)McCoy, Grant, Smolensky, Griffiths, and
  Linzen]{mccoy2020universal}
R~Thomas McCoy, Erin Grant, Paul Smolensky, Thomas~L Griffiths, and Tal Linzen.
\newblock Universal linguistic inductive biases via meta-learning.
\newblock \emph{arXiv preprint arXiv:2006.16324}, 2020.

\bibitem[M{\l}ynarski et~al.(2021)M{\l}ynarski, Hled{\'\i}k, Sokolowski, and
  Tka{\v{c}}ik]{mlynarski2021statistical}
Wiktor M{\l}ynarski, Michal Hled{\'\i}k, Thomas~R Sokolowski, and Ga{\v{s}}per
  Tka{\v{c}}ik.
\newblock Statistical analysis and optimality of neural systems.
\newblock \emph{Neuron}, 109\penalty0 (7):\penalty0 1227--1241, 2021.

\bibitem[Momennejad et~al.(2017)Momennejad, Russek, Cheong, Botvinick, Daw, and
  Gershman]{momennejad2017successor}
Ida Momennejad, Evan~M Russek, Jin~H Cheong, Matthew~M Botvinick,
  Nathaniel~Douglass Daw, and Samuel~J Gershman.
\newblock The successor representation in human reinforcement learning.
\newblock \emph{Nature human behaviour}, 1\penalty0 (9):\penalty0 680--692,
  2017.

\bibitem[Moss et~al.(2024)Moss, England, and Lio]{moss2024deep}
Jacob Moss, Jeremy England, and Pietro Lio.
\newblock Deep kernel learning of nonlinear latent force models.
\newblock \emph{Transactions on Machine Learning Research}, 2024.

\bibitem[Ober et~al.(2021)Ober, Rasmussen, and van~der Wilk]{ober2021promises}
Sebastian~W Ober, Carl~E Rasmussen, and Mark van~der Wilk.
\newblock The promises and pitfalls of deep kernel learning.
\newblock In \emph{Uncertainty in Artificial Intelligence}, pp.\  1206--1216.
  PMLR, 2021.

\bibitem[Ocko et~al.(2018)Ocko, Lindsey, Ganguli, and Deny]{ocko2018emergence}
Samuel Ocko, Jack Lindsey, Surya Ganguli, and Stephane Deny.
\newblock The emergence of multiple retinal cell types through efficient coding
  of natural movies.
\newblock \emph{Advances in Neural Information Processing Systems}, 31, 2018.

\bibitem[Olshausen \& Field(1996)Olshausen and Field]{olshausen1996emergence}
Bruno~A Olshausen and David~J Field.
\newblock Emergence of simple-cell receptive field properties by learning a
  sparse code for natural images.
\newblock \emph{Nature}, 381\penalty0 (6583):\penalty0 607--609, 1996.

\bibitem[Paszke(2019)]{paszke2019pytorch}
A~Paszke.
\newblock Pytorch: An imperative style, high-performance deep learning library.
\newblock \emph{arXiv preprint arXiv:1912.01703}, 2019.

\bibitem[Patacchiola et~al.(2020)Patacchiola, Turner, Crowley, O'Boyle, and
  Storkey]{patacchiola2020bayesian}
Massimiliano Patacchiola, Jack Turner, Elliot~J Crowley, Michael O'Boyle, and
  Amos~J Storkey.
\newblock Bayesian meta-learning for the few-shot setting via deep kernels.
\newblock \emph{Advances in Neural Information Processing Systems},
  33:\penalty0 16108--16118, 2020.

\bibitem[Pitkow \& Meister(2012)Pitkow and Meister]{pitkow2012decorrelation}
Xaq Pitkow and Markus Meister.
\newblock Decorrelation and efficient coding by retinal ganglion cells.
\newblock \emph{Nature neuroscience}, 15\penalty0 (4):\penalty0 628--635, 2012.

\bibitem[Qiu et~al.(2023{\natexlab{a}})Qiu, Klindt, Szatko, Gonschorek,
  Hoefling, Schubert, Busse, Bethge, and Euler]{qiu2023dataset}
Yongrong Qiu, David Klindt, Klaudia Szatko, Dominic Gonschorek, Larissa
  Hoefling, Timm Schubert, Laura Busse, Matthias Bethge, and Thomas Euler.
\newblock Efficient coding of natural scenes improves neural system
  identification (v1.0) [data set], 2023{\natexlab{a}}.
\newblock URL \url{https://doi.org/10.5281/zenodo.7656868}.

\bibitem[Qiu et~al.(2023{\natexlab{b}})Qiu, Klindt, Szatko, Gonschorek,
  Hoefling, Schubert, Busse, Bethge, and Euler]{qiu2023efficient}
Yongrong Qiu, David~A Klindt, Klaudia~P Szatko, Dominic Gonschorek, Larissa
  Hoefling, Timm Schubert, Laura Busse, Matthias Bethge, and Thomas Euler.
\newblock Efficient coding of natural scenes improves neural system
  identification.
\newblock \emph{PLoS computational biology}, 19\penalty0 (4):\penalty0
  e1011037, 2023{\natexlab{b}}.

\bibitem[Rahimi \& Recht(2007)Rahimi and Recht]{rahimi2007random}
Ali Rahimi and Benjamin Recht.
\newblock Random features for large-scale kernel machines.
\newblock \emph{Advances in neural information processing systems}, 20, 2007.

\bibitem[Rasmussen \& Ghahramani(2000)Rasmussen and
  Ghahramani]{rasmussen2000occam}
Carl Rasmussen and Zoubin Ghahramani.
\newblock Occam's razor.
\newblock \emph{Advances in neural information processing systems}, 13, 2000.

\bibitem[Ringach et~al.(2002)Ringach, Shapley, and
  Hawken]{ringach2002orientation}
Dario~L Ringach, Robert~M Shapley, and Michael~J Hawken.
\newblock Orientation selectivity in macaque v1: diversity and laminar
  dependence.
\newblock \emph{Journal of neuroscience}, 22\penalty0 (13):\penalty0
  5639--5651, 2002.

\bibitem[Rudi \& Rosasco(2017)Rudi and Rosasco]{rudi2017generalization}
Alessandro Rudi and Lorenzo Rosasco.
\newblock Generalization properties of learning with random features.
\newblock \emph{Advances in neural information processing systems}, 30, 2017.

\bibitem[Saxe et~al.(2011)Saxe, Koh, Chen, Bhand, Suresh, and
  Ng]{saxe2011random}
Andrew~M Saxe, Pang~Wei Koh, Zhenghao Chen, Maneesh Bhand, Bipin Suresh, and
  Andrew~Y Ng.
\newblock On random weights and unsupervised feature learning.
\newblock In \emph{Icml}, volume~2, pp.\ ~6, 2011.

\bibitem[Shrinivasan et~al.(2023)Shrinivasan, Lurz, Restivo, Denfield, Tolias,
  Walker, and Sinz]{shrinivasan2023taking}
Suhas Shrinivasan, Konstantin-Klemens Lurz, Kelli Restivo, George Denfield,
  Andreas Tolias, Edgar Walker, and Fabian Sinz.
\newblock Taking the neural sampling code very seriously: A data-driven
  approach for evaluating generative models of the visual system.
\newblock \emph{Advances in Neural Information Processing Systems},
  36:\penalty0 21945--21959, 2023.

\bibitem[Simoncelli \& Olshausen(2001)Simoncelli and
  Olshausen]{simoncelli2001natural}
Eero~P Simoncelli and Bruno~A Olshausen.
\newblock Natural image statistics and neural representation.
\newblock \emph{Annual review of neuroscience}, 24\penalty0 (1):\penalty0
  1193--1216, 2001.

\bibitem[Smirnakis et~al.(1997)Smirnakis, Berry, Warland, Bialek, and
  Meister]{smirnakis1997adaptation}
Stelios~M Smirnakis, Michael~J Berry, David~K Warland, William Bialek, and
  Markus Meister.
\newblock Adaptation of retinal processing to image contrast and spatial scale.
\newblock \emph{Nature}, 386\penalty0 (6620):\penalty0 69--73, 1997.

\bibitem[Standley et~al.(2020)Standley, Zamir, Chen, Guibas, Malik, and
  Savarese]{standley2020tasks}
Trevor Standley, Amir Zamir, Dawn Chen, Leonidas Guibas, Jitendra Malik, and
  Silvio Savarese.
\newblock Which tasks should be learned together in multi-task learning?
\newblock In \emph{International conference on machine learning}, pp.\
  9120--9132. PMLR, 2020.

\bibitem[Todorov \& Jordan(2002)Todorov and Jordan]{todorov2002optimal}
Emanuel Todorov and Michael~I Jordan.
\newblock Optimal feedback control as a theory of motor coordination.
\newblock \emph{Nature neuroscience}, 5\penalty0 (11):\penalty0 1226--1235,
  2002.

\bibitem[Ulyanov et~al.(2018)Ulyanov, Vedaldi, and Lempitsky]{ulyanov2018deep}
Dmitry Ulyanov, Andrea Vedaldi, and Victor Lempitsky.
\newblock Deep image prior.
\newblock In \emph{Proceedings of the IEEE conference on computer vision and
  pattern recognition}, pp.\  9446--9454, 2018.

\bibitem[Van~Amersfoort et~al.(2021)Van~Amersfoort, Smith, Jesson, Key, and
  Gal]{van2021feature}
Joost Van~Amersfoort, Lewis Smith, Andrew Jesson, Oscar Key, and Yarin Gal.
\newblock On feature collapse and deep kernel learning for single forward pass
  uncertainty.
\newblock \emph{arXiv preprint arXiv:2102.11409}, 2021.

\bibitem[Van~Hateren \& van~der Schaaf(1998)Van~Hateren and van~der
  Schaaf]{van1998independent}
J~Hans Van~Hateren and Arjen van~der Schaaf.
\newblock Independent component filters of natural images compared with simple
  cells in primary visual cortex.
\newblock \emph{Proceedings of the Royal Society of London. Series B:
  Biological Sciences}, 265\penalty0 (1394):\penalty0 359--366, 1998.

\bibitem[Vargas et~al.(2024)Vargas, Bisi, Chiappa, Versteeg, Miller, and
  Mathis]{vargas2024task}
Alessandro~Marin Vargas, Axel Bisi, Alberto~S Chiappa, Chris Versteeg, Lee~E
  Miller, and Alexander Mathis.
\newblock Task-driven neural network models predict neural dynamics of
  proprioception.
\newblock \emph{Cell}, 187\penalty0 (7):\penalty0 1745--1761, 2024.

\bibitem[Vershynin(2018)]{vershynin2018high}
Roman Vershynin.
\newblock \emph{High-dimensional probability: An introduction with applications
  in data science}, volume~47.
\newblock Cambridge university press, 2018.

\bibitem[Wilcoxon(1992)]{wilcoxon1992individual}
Frank Wilcoxon.
\newblock Individual comparisons by ranking methods.
\newblock In \emph{Breakthroughs in statistics: Methodology and distribution},
  pp.\  196--202. Springer, 1992.

\bibitem[Williams \& Rasmussen(2006)Williams and
  Rasmussen]{williams2006gaussian}
Christopher~KI Williams and Carl~Edward Rasmussen.
\newblock \emph{Gaussian processes for machine learning}, volume~2.
\newblock MIT press Cambridge, MA, 2006.

\bibitem[Wilson \& Adams(2013)Wilson and Adams]{wilson2013gaussian}
Andrew Wilson and Ryan Adams.
\newblock Gaussian process kernels for pattern discovery and extrapolation.
\newblock In \emph{International conference on machine learning}, pp.\
  1067--1075. PMLR, 2013.

\bibitem[Wilson \& Izmailov(2020)Wilson and Izmailov]{wilson2020bayesian}
Andrew~G Wilson and Pavel Izmailov.
\newblock Bayesian deep learning and a probabilistic perspective of
  generalization.
\newblock \emph{Advances in neural information processing systems},
  33:\penalty0 4697--4708, 2020.

\bibitem[Wilson et~al.(2016)Wilson, Hu, Salakhutdinov, and
  Xing]{wilson2016deep}
Andrew~Gordon Wilson, Zhiting Hu, Ruslan Salakhutdinov, and Eric~P Xing.
\newblock Deep kernel learning.
\newblock In \emph{Artificial intelligence and statistics}, pp.\  370--378.
  PMLR, 2016.

\bibitem[Yamins et~al.(2014)Yamins, Hong, Cadieu, Solomon, Seibert, and
  DiCarlo]{yamins2014performance}
Daniel~LK Yamins, Ha~Hong, Charles~F Cadieu, Ethan~A Solomon, Darren Seibert,
  and James~J DiCarlo.
\newblock Performance-optimized hierarchical models predict neural responses in
  higher visual cortex.
\newblock \emph{Proceedings of the national academy of sciences}, 111\penalty0
  (23):\penalty0 8619--8624, 2014.

\bibitem[Zagorski et~al.(2017)Zagorski, Tabata, Brandenberg, Lutolf,
  Tka{\v{c}}ik, Bollenbach, Briscoe, and Kicheva]{zagorski2017decoding}
Marcin Zagorski, Yoji Tabata, Nathalie Brandenberg, Matthias~P Lutolf,
  Ga{\v{s}}per Tka{\v{c}}ik, Tobias Bollenbach, James Briscoe, and Anna
  Kicheva.
\newblock Decoding of position in the developing neural tube from antiparallel
  morphogen gradients.
\newblock \emph{Science}, 356\penalty0 (6345):\penalty0 1379--1383, 2017.

\bibitem[Zhang \& Yang(2018)Zhang and Yang]{zhang2018overview}
Yu~Zhang and Qiang Yang.
\newblock An overview of multi-task learning.
\newblock \emph{National Science Review}, 5\penalty0 (1):\penalty0 30--43,
  2018.

\bibitem[Zhuang et~al.(2021)Zhuang, Yan, Nayebi, Schrimpf, Frank, DiCarlo, and
  Yamins]{zhuang2021unsupervised}
Chengxu Zhuang, Siming Yan, Aran Nayebi, Martin Schrimpf, Michael~C Frank,
  James~J DiCarlo, and Daniel~LK Yamins.
\newblock Unsupervised neural network models of the ventral visual stream.
\newblock \emph{Proceedings of the National Academy of Sciences}, 118\penalty0
  (3):\penalty0 e2014196118, 2021.

\end{thebibliography}
